%% file: main.tex
\documentclass[11pt]{article}

\usepackage{microtype}
\usepackage{graphicx}
\usepackage{subcaption}
\usepackage{booktabs} %

\usepackage{times}
\usepackage{latexsym}
\usepackage{inconsolata}

\usepackage{hyperref}
\usepackage{pifont}

\usepackage[utf8]{inputenc} %
\usepackage[T1]{fontenc}    %
\usepackage{hyperref}       %
\usepackage{url}            %
\usepackage{booktabs}       %
\usepackage{amsfonts}       %
\usepackage{nicefrac}       %
\usepackage{microtype}      %
\usepackage{xcolor}         %
\usepackage[many]{tcolorbox}
\usepackage{graphicx}
\usepackage{caption}
\usepackage{multirow}
\usepackage{subcaption}
\usepackage{enumitem}

\definecolor{researchercolor}{HTML}{7700B7}
\definecolor{reviewercolor}{HTML}{60BF00}
\definecolor{envfill}{HTML}{FFFFFF}

\newtcolorbox{researcherpromptbox}[1]{
    enhanced,
    breakable,
    boxrule=1pt,  %
    fontupper=\small,
    fonttitle=\bfseries\color{black},
    arc=3pt,  %
    rounded corners,
    colframe=researchercolor,
    colbacktitle=envfill,
    colback=envfill,
    title=#1,
    left=2mm,  %
    right=2mm,  %
    top=1mm,  %
    bottom=1mm  %
}

\newtcolorbox{reviewerpromptbox}[1]{
    enhanced,
    breakable,
    boxrule=1pt,  %
    fontupper=\small,
    fonttitle=\bfseries\color{black},
    arc=3pt,  %
    rounded corners,
    colframe=reviewercolor,
    colbacktitle=envfill,
    colback=envfill,
    title=#1,
    left=2mm,  %
    right=2mm,  %
    top=1mm,  %
    bottom=1mm  %
}

\usepackage{acl}

\usepackage{amsmath}
\usepackage{amssymb}
\usepackage{mathtools}
\usepackage{amsthm}

\usepackage[capitalize,noabbrev]{cleveref}

\theoremstyle{plain}

\theoremstyle{definition}

\theoremstyle{remark}

\ifdefined\hidecomments
  \newcommand\talor[1]{}
  \newcommand\gal[1]{}
\else
  \newcommand\talor[1]{{\color{brown}Talor: #1}}
  \newcommand\gal[1]{{\color{cyan}Gal: #1}}
\fi

\newcommand{\midtab}{\hspace{10pt}}
\newcommand{\benchmark}{AblationBench}
\newcommand{\researcherbench}{AuthorAblation}
\newcommand{\reviewerbench}{ReviewerAblation}
\newcommand{\judgedataset}{JudgeEval}
\newcommand{\researcherjudgedataset}{AuthorEval}
\newcommand{\reviewerjudgedataset}{ReviewerEval}

\usepackage[textsize=tiny]{todonotes}

\title{\benchmark{}: Evaluating Automated Planning of Ablations in Empirical AI Research}

\author{Talor Abramovich \\
  Tel Aviv University \\
  \texttt{talora1@mail.tau.ac.il} \\\And
  Gal Chechik \\
  Bar Ilan University \\
  NVIDIA}

\begin{document}
\maketitle

\begin{abstract}

Language model agents are increasingly used to automate scientific research, yet evaluating their scientific contributions remains a challenge.
A key mechanism to obtain such insights is through ablation experiments. To this end, we introduce \emph{\benchmark{}}, a benchmark suite for evaluating agents on ablation planning tasks in empirical AI research. It includes two tasks: \emph{\researcherbench{}}, which helps authors propose ablation experiments based on a method section and contains 83 instances, and \emph{\reviewerbench{}}, which helps reviewers find missing ablations in a full paper and contains 350 instances. For both tasks, we develop LM-based judges that serve as an automatic evaluation framework. Our experiments with frontier LMs show that these tasks remain challenging, with the best-performing LM system identifying only 45\% of the original ablations on average, below human-level performance.
We observe an inverse performance trend between the author and reviewer tasks, which we attribute to differences in model grounding.
Lastly, we analyze the limitations of current LMs on these tasks, and find that chain-of-thought prompting outperforms an agent-based approach.\footnote{Data: \href{https://huggingface.co/collections/ai-coscientist/ablationbench}{\raisebox{-0.25\height}{\includegraphics[height=3ex]{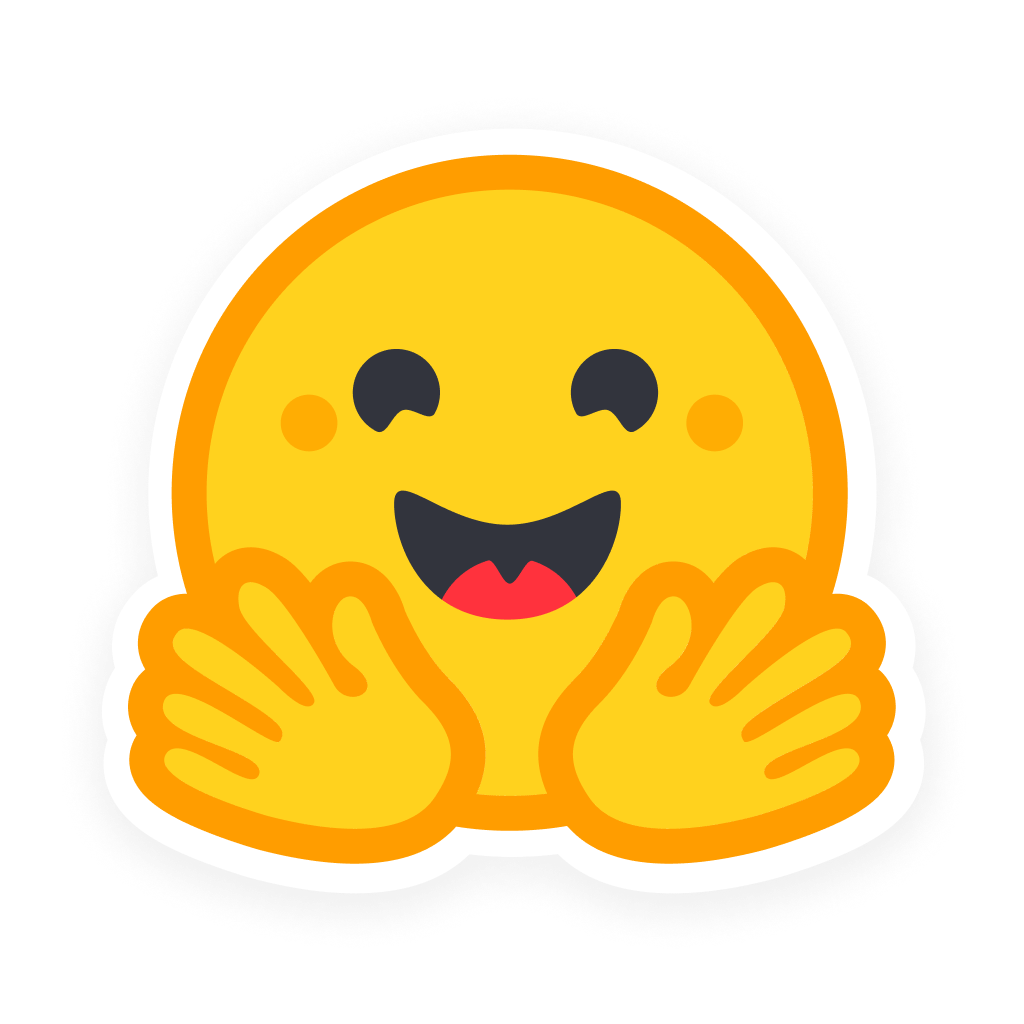}} HuggingFace}. Code: \href{https://github.com/ai-scientist-bench/ablation-bench}{\raisebox{-0.25\height}{\includegraphics[height=3ex]{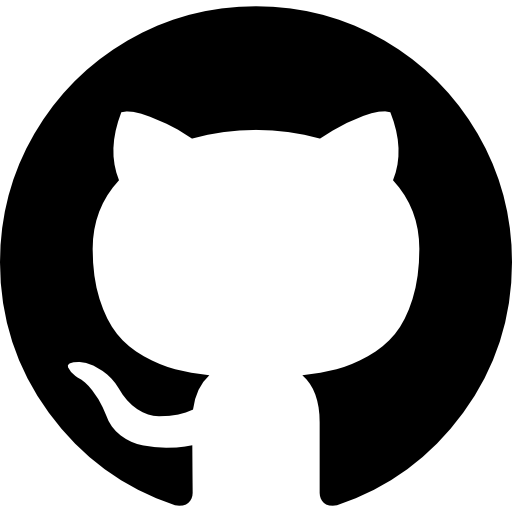}} GitHub}.}
\end{abstract}

\section{Introduction} \label{sec:introduction}

Automating scientific research using language model agents is a fast-growing field. Several groups have shown LM agents, often referred to as \emph{AI co-scientists},  that automate various steps of the scientific process from hypothesis generation \cite{lu2024aiscientistfullyautomated, si2024llmsgeneratenovelresearch, gottweis2025aicoscientist} to experiment execution \cite{bogin2024superevaluatingagentssetting, tian2024scicoderesearchcodingbenchmark, xiao2025csrbenchbenchmarkingllmagents, nathani2025mlgymnewframeworkbenchmark, starace2025paperbenchevaluatingaisability}. Evaluating the efficacy of these agents in generating insightful scientific contributions poses a significant challenge. A 
 key mechanism to obtain such insights is by dissecting a proposed method into its components. The contribution of these components is evaluated by \textit{ablation experiments} ~\cite{meyes2019ablationstudiesartificialneural, lipton2019troublingtrends}.

\begin{figure*}[t]
    \centering
    \includegraphics[width=\textwidth]{assets/figure1.pdf}
  \caption{Overview of AblationBench. (\textbf{Top}) \textbf{\researcherbench{}} is a collection of conference papers where the task of the planner is to generate ablation plan based on the paper's method section. The judge evaluates the generated plan using manually extracted gold ablations from the paper. (\textbf{Bottom}) \textbf{\reviewerbench{}} is a collection of ICLR submissions where the task of the planner is to generate missing ablation based on the full submission. The judge evaluates the generated ablations using the official reviews of the submission. %
  } \label{fig:1}
\end{figure*}

One can separate the process of ablation experiments into two steps: \textit{planning} these experiments, and \textit{executing} them by writing the code and collecting results. Here, we identify planning as a key step in AI co-scientists~\cite{wei2022chain, Wang2023PlanandSolvePI, yao2023tree, DBLP:journals/corr/abs-2402-02716}.
Other studies developed approaches for research code generation, e.g. \cite{tian2024scicoderesearchcodingbenchmark,xiao2025csrbenchbenchmarkingllmagents,starace2025paperbenchevaluatingaisability}, and this is not the focus of this paper. Planning ablations is challenging because it requires identifying key components of a method, and then proposing meaningful changes that are relevant, feasible, and grounded in domain knowledge.
An important first step toward automatic ablation planning was introduced in AbGen~\cite{zhao-etal-2025-abgen}. AbGen generates ablation plans from manually processed paper content that summarizes the research background, methodology, main experiment and results. An LM is then instructed to propose ablations for a specific component of the method. These ablations are evaluated mainly by human experts who rank their quality. This leaves fully automated ablation planning from raw papers unexplored.

Here, we look into two tasks in automating ablations: (1) Assisting \textbf{\textit{authors}}, proposing ablation experiments given a written method section; and (2) Assisting \textbf{\textit{reviewers}}: proposing missing ablation experiments of a given paper. 
We construct benchmark datasets for developing ablation agents for these two tasks. The first, \emph{\researcherbench{}} contains 83 papers, spans across 14 conferences and includes 230 human-annotated ablation experiments from the original papers. The second, \emph{\reviewerbench{}} is based on 350 ICLR submissions and their official reviews from 2023-2025 that mention ablations. Also, for each of the two tasks we develop an LM-based judge - an evaluation agent using ground truth data, that can evaluate any author-assistant or reviewer-assistant applied to our benchmarks.  Together they provide a complete evaluation resource for developing ablation agents.

We establish baseline levels of performance for our two ablation benchmark tasks  (author and reviewer) using two ``planners" (LM and Agent). These planners are LM systems that generate ablation plans. The best planner identifies at most 45\% of the original ablations across tasks. On a subset of ten \researcherbench{} tasks, humans achieve an F1 score of 0.65, compared to 0.42 for the best-performing model. These results highlight the benchmark's difficulty.

In summary this paper makes the following contributions: \\
\noindent (1) \emph{\benchmark{}}: A benchmark suite for evaluating LMs in ablation planning, containing two tasks:
\begin{itemize}[nosep,after=\strut]
\item \emph{\researcherbench{}}: A task of generating an ablation plan based on a paper's method section, consisting of 83 research papers with 230 human-annotated ablations. 
\item \emph{\reviewerbench{}}: A task of identifying missing ablations in paper submissions, consisting of 350 ICLR submissions and their official reviews. 
\end{itemize}
\noindent (2) Two LM-based judges for evaluating generated plans against ground truth for each task, together with a quantitative evaluation of the accuracy of the two judges using a held-out evaluation dataset with human annotations. 
~\newline\noindent (3) Two baseline LM planners for the benchmark to assess the capabilities of current frontier LMs on ablation planning.

\section{Related Work} \label{sec:related_work}

\paragraph{AI co-scientists.} 
Several groups have proposed automated pipelines for the scientific research cycle using AI agents~\cite{ifargan2024autonomousllmdrivenresearchdata, lu2024aiscientistfullyautomated, bubeck2025earlyscienceaccelerationexperiments, tang2025airesearcherautonomousscientificinnovation}.
\citet{gottweis2025aicoscientist} introduce a multi-agent system for hypothesis generation in biomedical research that assists human scientists, while \citet{audranreiss2025doesgoodairesearch} analyze the strengths and failure modes of such systems. Our work focuses on isolating and evaluating a single core scientific capability: ablation experiment design.

\paragraph{Co-scientists benchmarks.}
Several benchmarks evaluate AI systems across stages of scientific research.
~\citet{song2025evaluatinglargelanguagemodels} evaluates models on scientific discovery projects in various domains. CiteME~\cite{press2024citeme}, LitSearch~\cite{ajith2024litsearchretrievalbenchmarkscientific}, and PaperQA2~\cite{skarlinski2024languageagentsachievesuperhuman} focus on citation accuracy for claims. \citet{si2024llmsgeneratenovelresearch} evaluate AI research idea generation. Many benchmarks focus on experiment replication and code generation from papers~\cite{bogin2024superevaluatingagentssetting, huang2024mlagentbenchevaluatinglanguageagents, tian2024scicoderesearchcodingbenchmark, xiao2025csrbenchbenchmarkingllmagents, nathani2025mlgymnewframeworkbenchmark, starace2025paperbenchevaluatingaisability, siegel2024corebenchfosteringcredibilitypublished, seo2025paper2codeautomatingcodegeneration}, or solving scientific tasks like Kaggle challenges~\cite{chan2025mlebenchevaluatingmachinelearning, wijk2024rebenchevaluatingfrontierai, gu2024bladebenchmarkinglanguagemodel}. We build our benchmark on top of prior work on code replication, with a focus on ablation studies in empirical AI research.
AbGen~\cite{zhao-etal-2025-abgen} is a recent benchmark focused on ablation planning, where an LM is instructed to design ablation plan given a manually processed research context, method and main experiment from a single paper. \benchmark{} extends this line of work in several important ways. Most notably, it establishes a fully automated pipeline for ablation planning and evaluation, where models must infer full ablation plans directly from raw paper content, rather than from preprocessed or structured inputs.  Additionally, \benchmark{} formulates ablation planning from two complementary perspectives: authors and reviewers, covering multiple conferences and topics in AI research. Full comparison appears in~\cref{table:abgen_comparison}.

\begin{table}[tb]

\centering
\setlength{\tabcolsep}{2.5pt}
\small
\resizebox{\linewidth}{!}{
\begin{tabular}{l p{4cm} p{4cm}}
\toprule
 & \textbf{\benchmark{}} & \textbf{AbGen~\cite{zhao-etal-2025-abgen}} \\
\midrule

Task
& Author and reviewer
& Author only \\

Input
& Raw paper text
& Manually processed research context from a single paper  \\

Output
& Full ablation plan
& Specified component ablation \\

Evaluation
& LM-based judges
& Human \\

Criteria
& Compared to ground truth
& Subjective ablation quality \\

Domain
& Empirical AI (ML, CV, NLP)
& NLP \\

\# Papers
& 433
& 807 \\

\bottomrule
\end{tabular}
}
\caption{\benchmark{} and AbGen comparison.}
\label{table:abgen_comparison}
\end{table}

\paragraph{LM agents for scientific research.} 
LM agents for scientific tasks often build on ReAct~\cite{yao2023react}, where the model iteratively generates thoughts, performs actions, and receives observations in a containerized environment. Examples include MLAgentBench~\cite{huang2024mlagentbenchevaluatinglanguageagents}, AutoKaggle~\cite{li2024autokagglemultiagentframeworkautonomous}, and AgentK~\cite{grosnit2024largelanguagemodelsorchestrating}. More recently, MLGym~\cite{nathani2025mlgymnewframeworkbenchmark} introduced a framework based on SWE-agent~\cite{yang2024sweagent}, an agent built to automate software engineering using specialized agent-computer interfaces. MLGym is tailored for AI research tasks with built-in agent-computer interfaces for that matter. %

\paragraph{LMs as evaluation judges.}
Evaluating model-generated output is inherently ambiguous—different plausible outputs may vary in structure or wording, even if they capture the same underlying idea.
To address this, prior work has researched LMs-as-judges~\cite{Liu2023GEvalNE, zheng2023judging, Fu2023GPTScoreEA, Chan2023ChatEvalTB, zhuge2024agentasajudgeevaluateagentsagents}, often with chain-of-thought (CoT) prompting~\cite{wei2023chainofthoughtpromptingelicitsreasoning}, showing performance on par with or better than human evaluations~\cite{chiang-lee-2023-large, 10534765, ahmed2025llmsreplacemanualannotation}.
Despite their strong performance, LM-based judges can exhibit systematic biases, including positional, intra-model, and contextual biases~\cite{wang-etal-2024-large-language-models-fair, verga2024replacingjudgesjuriesevaluating, xu-etal-2024-pride, zhou2024batchcalibrationrethinkingcalibration}. Prior work has proposed mitigation strategies such as using diverse judge models or relying on pairwise comparisons instead of direct ranking. We follow this line of work to design our automatic evaluation system.

\section{Overview of \benchmark{}} 

\benchmark{} is designed for two tasks: First, for generating ablation plans based on a paper's method section, we built a component called \textit{\researcherbench{}}.  Second, to identify missing ablations in a full paper, similar to what a reviewer would do, we built  \textit{\reviewerbench{}}. 

For each task, we start by collecting a set of papers for the task setup. Next, we design \textit{LM planners} to generate ablation plans for each of these two tasks. Finally, we develop \textit{LM judges} to evaluate the quality of the generated plans by comparing them to gold labels. The judges are used to automate the evaluation of our benchmark, inspired by the judge used in PaperBench~\cite{starace2025paperbenchevaluatingaisability}.
In both tasks, the judges evaluate against a ground-truth (GT): the ablations originally conducted by the authors in \researcherbench{}, and the ablations suggested by reviewers in \reviewerbench{}. While models may propose new ablation ideas, evaluating such suggestions is challenging, as no GT exists for this setting, and the suggestions must be assessed for both relevance and feasibility, so we leave this as future work. With that said, we argue in the next sections that the ablations proposed by the authors and reviewers provide a meaningful evaluation target, and we stress that current models still struggle with finding those ablations.

To evaluate how well the judges perform, we also construct a dedicated dataset, \textit{\judgedataset{}}, consisting of \textit{\researcherjudgedataset{}} and \textit{\reviewerjudgedataset{}}, with manually labeled examples.

We report the overall \benchmark{} score as the average performance across the two tasks, for a complete view of both the author and reviewer perspectives in ablation planning.
\Cref{fig:1} shows an overview of \benchmark{}. We now describe each component in detail.

\section{\researcherbench{}: Ablation Benchmark for Authors} \label{sec:researcher_benchmark}
The first \benchmark{} component is designed for the task of proposing ablation experiments given a paper including its method section but without the experiments sections. We collect 83 papers and annotate each with the ablations that appear in the full paper, serving as gold labels.

We develop two baseline planners for this task. For a given instance they each generate $k$ proposals for ablation experiments. We also design an automatic judge to evaluate planners. To assess the quality of the judge itself, we created a separate dataset of 63 plans that is human annotated, so we can compare the decisions of the judge with human evaluation of the ablation proposals. 

In the following, we describe the construction, planner, and judge for \researcherbench{} component.

\subsection{Constructing \researcherbench{}} \label{sec:researcher_bench_construction}

\researcherbench{} is built from 83 machine learning conference papers that include ablation studies and open-source code. We collect these papers from prior benchmarks—CSR-Bench (100 papers)~\cite{xiao2025csrbenchbenchmarkingllmagents}, SUPER-Expert (45 papers)~\cite{bogin2024superevaluatingagentssetting}, and PaperBench (20 papers)~\cite{starace2025paperbenchevaluatingaisability}, and supplement them with 38 best paper awardees from CVPR and ICCV (2020–2024) to increase topic and conference diversity.

We filter papers by checking for the presence of ablation sections, then truncate each to include only content up to the method section. The ablation plan from the full paper is extracted into a structured JSON format to serve as GT. We extract ablations in the order they appear in the paper, using this order as a proxy for their relative importance. Papers are split into 21 for development and 62 for testing. See~\Cref{app:researcher_bench_construction} for full construction details.

Since ablation planning is open-ended, there may be many valid and useful ablations that do not appear in the original paper. We further validate the usefulness of author ablations on the development set. We find that among 105 ablation generations of the best model (GPT 5.4), 58 matched the GT (55\%), while only 7 ablations (7\%) were valuable ablations beyond the GT.

\subsection{\researcherbench{} Planner}  \label{sec:researcher_bench_planner}
The planner in \researcherbench{} is tasked with generating an ablation plan given a paper's title, abstract, and truncated content up to the method section. The output is a structured JSONL file containing up to $k$ ablation entries. Each entry should describe the removal or modification of a  component of the proposed method, aiming to assess its contribution.

We implement two planner variants as baselines. The first, \textsc{LM-Planner}, is using a single CoT style prompt to an LM for generating ablation plan. The LM receives the title, abstract, and aggregated \TeX{} source of the paper, and outputs $k$ ablations along with its reasoning.

The second, \textsc{Agent-Planner}, is an agent-based planner built on SWE-agent~\cite{yang2024sweagent}, a ReAct-style LM agent. This planner operates in a containerized environment where it has access to the paper files and basic tools for file inspection and command execution. The agent is prompted similarly but can take intermediate steps, perform actions on the environment, and refine its output before submitting the final ablation plan.

\subsection{\researcherbench{} Judge} \label{sec:researcher_bench_judge}

We evaluate the quality of ablation plans by checking whether each GT ablation is correctly predicted by the model. For each GT ablation, we assess whether there exists at least one matching ablation in the generated plan. 
A match means that the proposed ablation captures the same idea as the GT one. This includes targeting the same part of the method and applying a similar type of change.
\Cref{fig:researcher_bench_match_criteria} in~\Cref{app:researcher_bench_construction} provides examples of the matching criteria. 

We report three metrics for this binary classification task. First, \textit{precision@$k$} measures how many of the top $k$ proposed ablations appear in the GT. Second, \textit{recall@$k$} measures how many GT ablations are correctly predicted. Lastly, we report the \textit{F1-score@$k$} which balances between the two. 

To support automatic evaluation, we design an LM-based judge that compares the GT ablations to the generated ones and decides whether each has a match. We implement two scaffoldings for the LM-based judge: (1) \textsc{LMJudge}, a CoT-style prompt to an LM; and (2) \textsc{AgentJudge}, a ReAct-based agent allowing multiple steps of reasoning and interaction with an environment before generating the final decision. We address three sources of potential bias. First, \textit{intra-model bias}, where a model may favor its own outputs, is addressed by using an ensemble of three models and taking a majority vote~\cite{verga2024replacingjudgesjuriesevaluating}. Second, \textit{contextual bias}, where a model may prefer the GT over the plan due to contextual cues, is mitigated by removing GT and plan tags and randomly assigning them as Side A and Side B, with the model matching in both directions~\cite{zhou2024batchcalibrationrethinkingcalibration}. Lastly, \textit{positional bias}, where a model may favor ablations based on their order, is mitigated by randomly shuffling ablations on both sides, with each model observing a different permutation~\cite{wang-etal-2024-large-language-models-fair}.

To evaluate judges we create a dataset,  \textit{\researcherjudgedataset{}}, of ablation plans for papers in the development set with correct match labels.

\section{\reviewerbench{}: Ablation Benchmark for Reviewers} \label{sec:reviewer_bench}

The second \benchmark{} component focuses on suggesting missing ablations for a full paper submission. The goal is to identify important ablations that were not reported but should have been. We collect 350 ICLR submissions along with their official reviews, serving as the gold labels. 

We develop two baseline planners for this task. Given a paper submission they generate $k$ proposals for missing ablations. We also design an automatic judge to evaluate planners. To assess the quality of the judge itself, we created a separate dataset of 60 plans that is human annotated, so we can compare the decisions of the judge with human evaluation of the ablation proposals. 
In the following, we describe the benchmark, planner, and judge for \reviewerbench{}.

\subsection{Constructing \reviewerbench{} } \label{sec:reviewer_bench_construction}
\reviewerbench{} is built from ICLR submissions from 2023 to 2025, where reviewers suggested missing ablations. This setting enables the development of reviewer-assistive models that can detect gaps in a paper's empirical validation.

We apply a three-stage pipeline: (1) collect $\sim$89,100 reviews from $\sim$22,800 OpenReview~\cite{openreview} submission forums and filter for reviews mentioning “ablat,”; (2) use a CoT-prompted LM to identify reviews with concrete suggestions for new ablations; and (3) retrieve the submitted PDFs and convert them to markdown for standardized model input. This automatic pipeline makes it easier to extend the benchmark to other conferences and years.
This process results in 5,960 papers in \reviewerbench{}. We randomly sample 50 papers for development and 300 for testing. Full construction details are provided in~\Cref{app:reviewer_bench_construction}.

Since reviewer suggestions may be inconsistent~\cite{cortes2021inconsistencyconferencepeerreview, neurips2021blog}, we further validate the usefulness of suggested ablations on the development set. We find that for 41 out of 50 papers (82\%), the authors accepted the suggested ablations as valid and incorporated them either in the rebuttal or in later revisions of the paper.

\subsection{\reviewerbench{} Planner} \label{sec:reviewer_bench_planner}

The \reviewerbench{} planner is given a paper's title, abstract, and full preprint, and is tasked with generating $k$ missing ablations—experiments that were not reported in the paper but should have been included. This simulates a reviewer's perspective, where the goal is to identify important omissions in the empirical evaluation.

We implement two planner variants as baselines for this task, both follow the same setup described in \Cref{sec:researcher_bench_planner}: the LM planner uses a single CoT-style prompt to generate up to $k$ missing ablations, while the agent planner uses a ReAct-style agent to iteratively construct its suggestions.

\subsection{\reviewerbench{} Judge}  \label{sec:reviewer_bench_judge}
We evaluate the quality of generated missing ablations by checking whether they match the suggested ablations by reviewers. For each generated ablation, we assess whether there is a matching ablation in one of the reviewer's comments.
A match means that the generated ablation captures the same underlying idea as the reviewer's suggestion. The ablated component must align in both, but specific actions (e.g., remove, replace) or precise modification details are only required when they are explicitly stated in the review.

We report three metrics for this binary classification task: \textit{precision@$k$}, which measures how many of the top $k$ generated ablations match the GT; \textit{recall@$k$}, which measures how many of the GT reviewer suggestions are correctly predicted; and the \textit{F1-score@$k$}, which balances the two.

To enable automatic evaluation, we use two LM-based judge architectures similar to the \researcherbench{} Judge introduced in~\Cref{sec:researcher_bench_judge}. Each judge compares a generated ablation to the reviews' suggestions and decides whether it has a match. We also address potential biases in a similar way: we use an ensemble of three judges to avoid intra-model bias and we randomly shuffle the order of ablations and reviews to reduce positional bias, so that each judge sees a different permutation.

To evaluate judges we construct a dataset, \textit{\reviewerjudgedataset{}}, of ablation plans for papers in the development set, human annotated with match labels.

\section{\benchmark{} Properties}
\benchmark{}, composed of \researcherbench{} and \reviewerbench{}, provides a diverse evaluation suite for ablation planning in empirical AI research. \researcherbench{} contains 230 ground-truth ablations (4 per paper on average), including both modification (59\%) and removal (41\%) ablations, with modification ablations often admitting multiple valid alternatives. \reviewerbench{} contains on average 2 reviewer-suggested ablations per paper, with nearly half of all ICLR 2023--2025 submissions including at least one review mentioning ablations, highlighting their central role in the review process. The benchmark spans 14 conferences and a broad range of AI topics, including general machine learning, NLP, computer vision, representation learning, generative models, and AI applications, enabling evaluation across diverse research domains. See~\Cref{app:ablationbench_props} for full details.

\section{Experiments and Results} \label{sec:experiments}

We first describe the experimental setup for the judges and the planners in \researcherbench{} and \reviewerbench{}. We then present the results of the judges and analyze their results. Finally, we report and analyze the main results of the baseline planners across both tasks, together with a human evaluation study performed on \researcherbench{}. Additional experiments and results are presented in~\Cref{app:ext_results}, and full prompts are provided in~\Cref{app:prompts}.

\subsection{Experimental Setup: Judges}

\paragraph{Agent judges.} 

For the agent judges, we use SWE-agent~\cite{yang2024sweagent} with several frontier LMs: GPT-5.4~\cite{singh2025openaigpt5card}, o3-mini~\cite{openai2025o3mini} (high reasoning), GPT~OSS~120B~\cite{openai2025gptoss120bgptoss20bmodel}, MiniMax~M2.5~\cite{minimax2026m25} and Qwen~3.5~122B~\cite{qwenteam2026qwen35omnitechnicalreport}. The agents use tools only for file creation and editing, and are also provided a tool to generate the submission file in the required format, with a submit command to verify evaluation formatting. Memory is managed using SWE-agent's default context handling, which keeps the last five observations and collapses earlier ones. Each agent is supplied with a demonstration trajectory illustrating how to perform matching and use the interface, following in-context learning~\cite{brown2020languagemodelsfewshotlearners}. Prompts are configured per benchmark's task, based on the matching criteria  in~\Cref{sec:researcher_bench_judge,sec:reviewer_bench_judge}, to create separate judges for each task. 

\paragraph{LM judges.}
For the LM judges, we use the same LMs as in the agent judges. Each LM is prompted to output a decision for every ablation, along with a reasoning field explaining its choice. The judges are  also provided with an example illustrating how to perform the matching, similar to the demonstration used for the agent judges.

\paragraph{Metrics.}
We report precision, recall, and F1-score for the judges. We also report the Cohen's kappa $(\kappa)$ agreement between human annotations and our judges' annotations. Evaluation is done on \researcherjudgedataset{} and \reviewerjudgedataset{} for the \researcherbench{} judge and \reviewerbench{} judge, respectively. The overall \judgedataset{} score is calculated as the average performance across the two tasks.

\subsection{Experimental Setup: Planners}

\begin{table}[t]
    \centering
    \setlength{\tabcolsep}{2pt}
    
    \centering

    \resizebox{0.9\linewidth}{!}{
    \small
    \begin{tabular}{lcccc}
    \toprule
        & Precision & Recall & F1 & $\kappa$ \\
    \midrule
        \textsc{LMJudge} & & & & \\
        \midtab GPT-5.4 & 0.77 & 0.81 & 0.76 & 0.56 \\
        \midtab o3-mini & 0.76 & \textbf{0.86} & 0.78 & 0.58 \\
        \midtab GPT-OSS-120B & 0.70 & 0.75 & 0.63 & 0.45 \\
        \midtab MiniMax-M2.5 & 0.74 & 0.80 & 0.73 & 0.51 \\
        \midtab Qwen3.5-122B-A10B & \textbf{0.79} & 0.81 & 0.77 & 0.57 \\
        \midtab Majority Vote & 0.78 & 0.84 & \textbf{0.79} & \textbf{0.62} \\
        \cmidrule(ll){1-5}
        \textsc{AgentJudge} & & & & \\
        \midtab GPT-5.4 & 0.70 & 0.74 & 0.69 & 0.45 \\
        \midtab o3-mini & \textbf{0.78} & 0.78 & 0.74 & 0.53 \\
        \midtab GPT-OSS-120B & 0.71 & 0.75 & 0.70 & 0.49 \\
        \midtab MiniMax-M2.5 & 0.74 & \textbf{0.80} & 0.73 & 0.49 \\
        \midtab Qwen3.5-122B-A10B & 0.73 & 0.77 & 0.72 & 0.48 \\
        \midtab Majority Vote & \textbf{0.78} & 0.79 & \textbf{0.76} & \textbf{0.57} \\
    \bottomrule
    \end{tabular}
    }
    \caption{\textbf{Evaluating judges:} Macro-average results of the two judge scaffoldings for the \judgedataset{}. The majority vote is based on the three best models (in terms of $\kappa$) used in each scaffolding.}
    \label{table:judge_results}
\end{table}

\paragraph{Number of generated ablations ($k$).}
We set $k = 5$ for \researcherbench{} and $k = 2$ for \reviewerbench{}, based on the benchmarks statistics presented in~\Cref{table:dataset_stats}.

\paragraph{Agent planners.}
For the agent planners, we use SWE-agent with several frontier LMs: GPT-5.4, Llama 3.1 405B Instruct~\cite{meta2024llama31}, GPT~OSS~120B, MiniMax~M2.5 and Qwen~3.5~122B. SWE-agent is configured with default tools for file inspection, editing, and command execution. We provide it with a command to submit the ablation plan and verify that it matches the required format. Prompts are tailored to both benchmark task, as described in~\Cref{sec:researcher_bench_planner,sec:reviewer_bench_planner}, resulting in two planners.

\paragraph{LM planners.}
For the LM planners, we use the same LMs as in the agent planners, with two additional smaller, reasoning-oriented models: o3-mini and Gemini 2.5 Flash~\cite{google2025gemini25}. Each LM is prompted to output an ablation plan based on the task definition, along with a reasoning field that explains its planning process.

\paragraph{Metrics.}
We report macro-average precision@$k$, recall@$k$, and F1-score@$k$. \benchmark{} score is calculated as the average metrics of the two tasks.

\subsection{Experimental Results: Judges} \label{sec:judge_results}

\Cref{table:judge_results} reports results for the judges on \judgedataset{}. \textsc{AgentJudge} does not outperform \textsc{LMJudge}. This suggests that in short contexts, a single CoT step works better than multi-step reasoning.
Below, we report agreement with human annotations and provide an error analysis to support this choice.

\paragraph{Majority vote better aligns with human annotations.}  
We build a majority-vote ensemble using all three model sets per scaffolding. We then select the best ensemble by $\kappa$. For \textsc{LMJudge}, the best ensemble uses GPT-5.4, o3-mini, and MiniMax~M2.5. For \textsc{AgentJudge}, it uses o3-mini, GPT~OSS~120B, and MiniMax~M2.5.
These ensembles also achieve the best classification scores. We adopt majority voting as it mitigates potential self-model biases and achieves better performance than the best individual LM, consistent with prior work~\cite{verga2024replacingjudgesjuriesevaluating}.

\begin{table}[t]
    \centering
    \setlength{\tabcolsep}{2pt}
    
    \centering

    \resizebox{0.94\linewidth}{!}{
    \small
    \begin{tabular}{lccc}
    \toprule
        & Precision@$k$ & Recall@$k$ & F1@$k$  \\
    \midrule
        \textsc{LM-Planner} & & & \\
        \midtab GPT-5.4 & \textbf{0.33} & \textbf{0.45} & \textbf{0.35} \\
        \midtab o3-mini & 0.26 & 0.33 & 0.27 \\
        \midtab Gemini 2.5 Flash & 0.29 & 0.35 & 0.29 \\
        \midtab Llama 3.1 405B Instruct & 0.24 & 0.33 & 0.26 \\
        \midtab GPT-OSS-120B & 0.28 & 0.40 & 0.31 \\
        \midtab MiniMax-M2.5 & 0.29 & 0.37 & 0.30 \\
        \midtab Qwen3.5-122B-A10B & 0.28 & 0.39 & 0.30 \\
        \cmidrule(ll){1-4}
        \textsc{Agent-Planner} & & & \\
        \midtab GPT-5.4 & \textbf{0.33} & \textbf{0.44} & \textbf{0.35} \\
        \midtab Llama 3.1 405B Instruct & 0.24 & 0.26 & 0.23 \\
        \midtab GPT-OSS-120B & 0.16 & 0.20 & 0.17 \\
        \midtab MiniMax-M2.5 & 0.18 & 0.24 & 0.19 \\
        \midtab Qwen3.5-122B-A10B & 0.29 & 0.37 & 0.30 \\
    \bottomrule
    \end{tabular}
    }
    \caption{\textbf{Evaluating planners:} Macro-average results of the two planner scaffoldings for the \benchmark{}.}
    \label{table:main_results}
\end{table}

\paragraph{Impact of inputs and models on judge performance.}
\Cref{table:error_researcher_judge_analysis,table:error_reviewer_judge_analysis} (\Cref{app:ext_results}) analyzes errors of the best and worst \textsc{LMJudge} models on \judgedataset{} (o3-mini vs. GPT~OSS~120B).
On \researcherbench{}, o3-mini mainly makes false negatives (51\% of its errors). GPT~OSS shows more surface matches (30\%), suggesting reliance on lexical cues.
On \reviewerbench{}, o3-mini often misses ablations (34\%). GPT~OSS shows higher rates (38\%), indicating sensitivity to ambiguous input.
These trends explain the overall results (\Cref{table:researcher_judge_results,table:reviewer_judge_results}): performance is higher on \researcherjudgedataset{} due to structured inputs, and lower on \reviewerjudgedataset{} due to unstructured review text.

\begin{figure}[tb]
    \centering
    \includegraphics[width=0.9\linewidth]{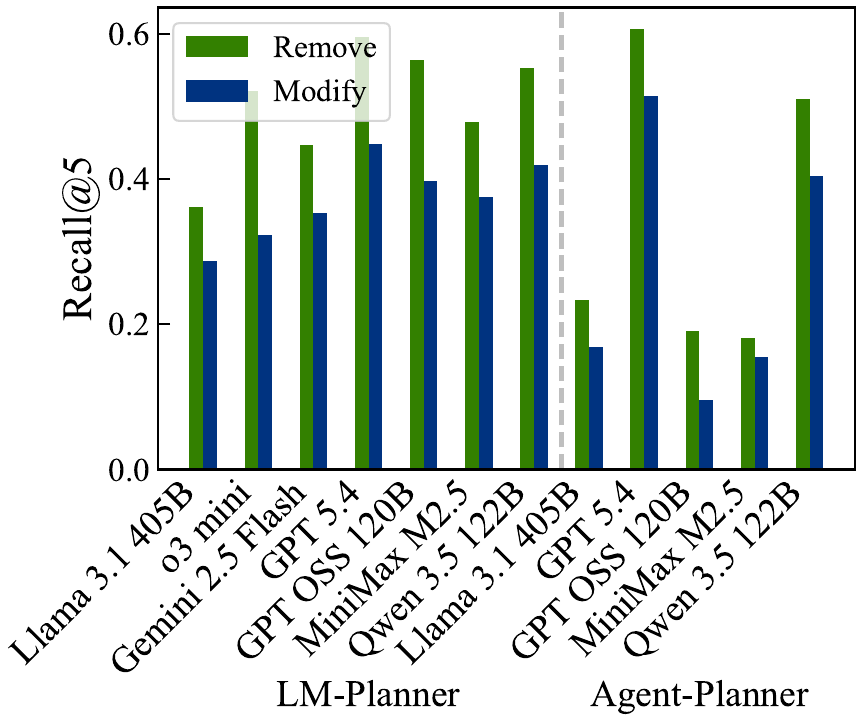}
    \caption{Recall@5 performance for \researcherbench{} of each planner approach across different LMs, separated by ablation type.}
    \label{fig:action_comparison}
\end{figure}

\subsection{Experimental Results: Planners} \label{sec:results}

\Cref{table:main_results} presents the main results for \benchmark{}. 
Below, we analyze the performance of the planners, highlighting where they succeed and what limits their current capabilities.

\paragraph{Models are more likely to identify ablations that completely remove key components of the method, rather than those that modify them.}
In~\Cref{fig:action_comparison} we observe that recall@5 is substantially higher for removal-type ablations across planners and models, suggesting that LMs are more effective at identifying core components of a method. In contrast, modification-type ablations require not only identifying the component but also proposing relevant and feasible substitutions, which remains a challenge for current models. Our qualitative analysis (\cref{app:ext_results}) shows that modification type ablations across models often emphasizes hyperparameter variations rather than suggesting meaningful replacements for core components. Models also tend to produce a fixed number of ablations, without prioritizing the most informative ones.

\begin{figure}[tb]
    \centering
    \includegraphics[width=0.93\linewidth]{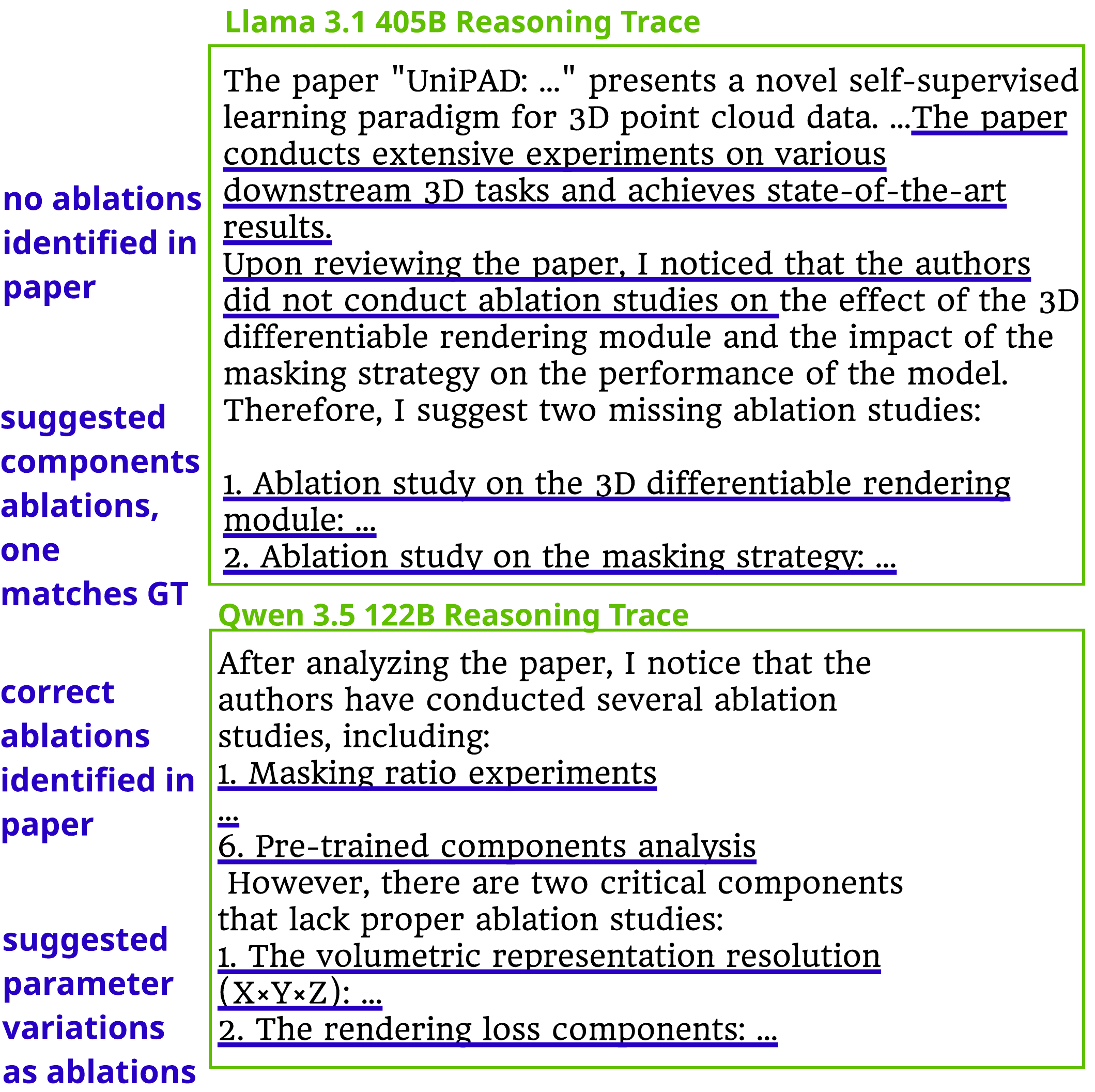}
  \caption{Qualitative analysis highlighting differences in the reasoning and grounding of Llama 3.1 405B and Qwen 3.5 122B on a \reviewerbench{} instance.} \label{fig:reasoning_analysis}
\end{figure}

\paragraph{Model grounding impacts performance across tasks.}
We qualitatively analyze ten randomly selected test papers, five per task, using the best- and worst-performing models, extracted from the results per each task. We observe a trade-off: models that perform well on one task tend to perform poorly on the other. Our analysis suggests that this behavior is associated with differences in model grounding.
More strongly grounded models, such as Qwen, perform well on \researcherbench{} by closely adhering to the method description and generating instruction-compliant ablation plans. However, this grounding appears to limit performance on \reviewerbench{} as illustrated in~\Cref{fig:reasoning_analysis}, where identifying missing ablations often requires creativity beyond the experiments explicitly described in the paper. In contrast, less grounded models such as Llama overlook contextual details, which hurts performance on \researcherbench{}, but can be advantageous for \reviewerbench{}. This observation aligns with prior findings on hallucination and factuality~\cite{Bang2025HalluLensLH, jacovi2025factsgroundingleaderboardbenchmarking}, particularly in long-context settings, which characterize both tasks.
 See~\Cref{app:ext_results} for a detailed analysis.

\paragraph{Performance varies by conference domain.}
When comparing performance across conference categories—General ML, CV, and NLP—we observe that F1 score is higher on CV papers and lower on NLP papers (0.45 vs. 0.30). Our qualitative analysis suggests this gap is due to differences in paper structure: CV papers typically follow a more standardized format with clearly defined components and ablations derived directly from this decomposition, while NLP papers are more variable and often less explicitly decompose components, making ablation planning harder (\Cref{fig:metrics_per_conf_type}).

\paragraph{\textsc{LM-Planner} vs. \textsc{Agent-Planner}.}
Overall, \textsc{LM-Planner} achieves slightly better classification metrics than \textsc{Agent-Planner}, outperforming it for Llama 3.1, GPT~OSS, and MiniMax-M2.5. For GPT-5.4, the best-performing model, \textsc{LM-Planner} attains an F1 score on par with \textsc{Agent-Planner}. We also observe no correlation between agent trajectory length and F1 score (Spearman: $0.07$). These results suggest that, despite their ability to iterate over files and perform multi-step reasoning, the agent scaffold used does not provide a clear advantage over a single LM call with full context and one CoT step. To test whether this finding depends on the specific scaffolding used, we additionally evaluate GPTResearcher~\cite{Elovic_gpt-researcher_2023} on AuthorAblation using its default configuration, while restricting retrieval from external sources containing the paper name to avoid contamination. GPTResearcher  achieves an F1@5 of 0.24, further supporting our finding that current agent scaffolds do not improve performance on this task.

\subsection{Human Baseline Performance} \label{sec:human_eval}

We establish a human baseline for \researcherbench{} using 10 PhD students in machine learning, each selecting one of 15 randomly sampled papers and generating ablation plans from truncated versions of the papers (details in~\Cref{app:human_evaluation_settings}). Humans achieve substantially higher performance than GPT-5.4, with F1@5 of 0.66 vs.\ 0.43 and $\times1.82$ precision, indicating a more selective and accurate identification of ablations. These results support the validity of the GT ablations and show that \benchmark{} remains a challenging, non saturated task with a clear gap for improvement.

\section{Conclusion} \label{sec:conclusion}

We introduced \textbf{\benchmark{}}, a suite for evaluating ablation planning in empirical AI research. It includes \textbf{\researcherbench{}}, which tasks models with generating ablations from a paper's method, and \textbf{\reviewerbench{}}, which focuses on suggesting missing ablations from full submissions—capturing both author and reviewer perspectives.
Each benchmark includes a baseline planner, and an LM judge for automatic evaluation of planners. CoT prompting outperforms agent-based methods in both accuracy and cost. The best model achieve an F1 score of 0.35 on \benchmark{}, leaving substantial room for improvement.

\section*{Limitations} 

Our current evaluation framework checks whether models generate ablations found in the paper or suggested in reviews. However, it has some limitations in the following.

\paragraph{LM-based judges achieve moderate human-model agreement.} Our automatic evaluation framework relies on LM-based judges, which achieve strong classification performance but only moderate agreement with human annotators. While this level of performance is sufficient for reliable evaluation, improving human–model agreement could further increase the stability and robustness of the judges.

\paragraph{GT covers only ablations proposed by authors and reviewers.} Models may generate valid and feasible ablations that are not present in the GT. We stress throughout this work that our GT as evaluation target is meaningful and that current models struggle in the task defined by \benchmark{}. Our qualitative analysis in~\Cref{app:ext_results} shows that, in most cases, GT ablations are preferred over those proposed by models, as model-generated plans often focus on hyperparameter variations rather than core methodological ablations. However, as models become stronger, they are expected to produce more meaningful ablation designs that go beyond the current GT, and such capabilities will be important to benchmark once \benchmark{} becomes saturated.

\paragraph{Potential data contamination.} \benchmark{} uses published conference papers, ICLR submissions and their official reviews, which frontier LMs may have seen during training. However, current models still struggle substantially with the task, suggesting that memorization is not a major driver of performance. Further, across the six models with disclosed knowledge cutoffs, F1 on instances published before vs. after the cutoff has only a weak correlation (Pearson: 0.35). While contamination may become more relevant on newer models, our fully automated data construction pipeline makes it easy to add newly published papers and construct temporally held-out evaluations for future models.

We hope \benchmark{} will support future research toward developing stronger models capable of creative ablation planning.

\bibliography{main}

\clearpage
\appendix
\crefalias{section}{appendix}
\crefalias{subsection}{appendix}
\crefalias{subsubsection}{appendix}

\newpage

\input{appendix}

\end{document}

%% file: appendix.tex
\section{\benchmark{} Construction}

\subsection{Constructing \researcherbench{}} \label{app:researcher_bench_construction}

We construct the \researcherbench{} benchmark using papers from previous datasets designed for scientific reasoning over machine learning papers. Specifically, we aggregate papers from CSR-Bench~\cite{xiao2025csrbenchbenchmarkingllmagents} (100 papers), SUPER-Expert~\cite{bogin2024superevaluatingagentssetting} (45 papers), and PaperBench~\cite{starace2025paperbenchevaluatingaisability} (20 papers). These sources were selected due to their size, high-quality paper selection process, availability of code, diversity of conferences and recency. 

To further increase topical and conference diversity, we supplement these with best paper awardees from major computer vision conferences (CVPR, ICCV) from 2020 to 2024 (38 papers). 

All selected papers should satisfy two criteria: they should include ablation studies and should have publicly released code, enabling future extensions of the benchmark to code development and execution settings.

\begin{figure*}[tb]
    \centering
    \includegraphics[width=0.98\linewidth]{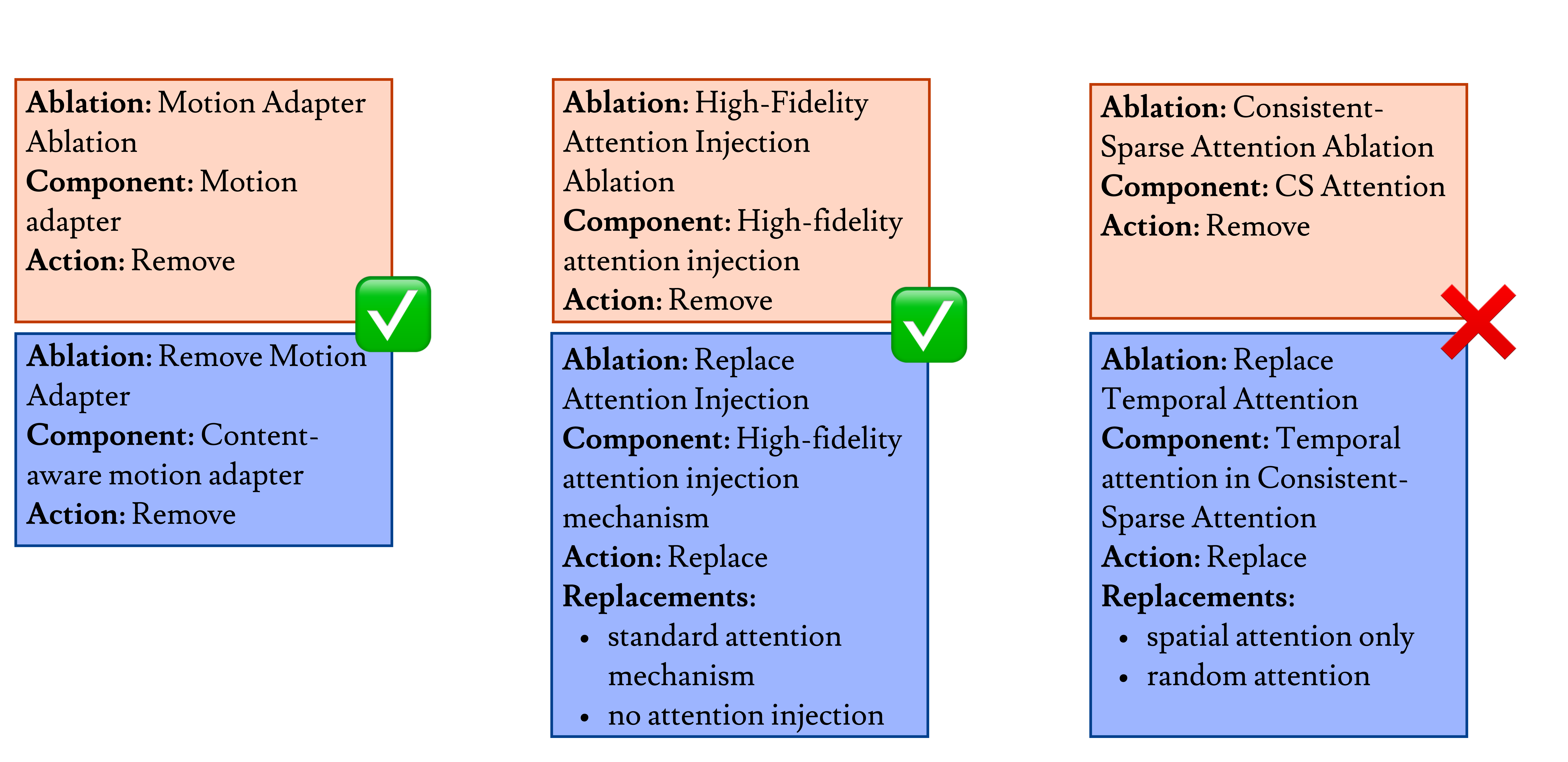}
  \caption{An example of the matching criteria in \researcherbench{} instance, showing alignment between GT (gold, top) and generated (blue, bottom) ablations. Matching is done semantically, and is represented as a bipartite graph: nodes correspond to GT and generated ablations and edges indicate valid matches. Evaluation metrics are computed with respect to the GT ablations; in this example, both precision and recall equal 0.75.} \label{fig:researcher_bench_match_criteria}
\end{figure*}

Based on these criteria we curate the benchmark using a three-stage pipeline:
\paragraph{1. Filtering for papers with ablations.} We collect the  \TeX{} source of all candidate papers and discard those that do not include ablation studies. This filtering is performed via keyword search ("ablat") and manual validation.
\paragraph{2. Partial paper creation.} For each remaining paper, we create a truncated version containing only the sections up to and including the method. To ensure that only content relevant to the model's task is included, we apply the arXiv \LaTeX{} cleaner utility~\cite{arxivlatexcleaner} to remove comments and any unreferenced files, such as figures or tables associated with later sections. This setup ensures that the model generates ablations based solely on the method description, without access to experimental results or analysis.
\paragraph{3. Ground-truth (GT) extraction.} The ablation plan described in the full paper is manually extracted into a structured JSON format, which serves as the GT output for evaluation. The instructions for annotators are given in~\Cref{fig:annotator_instructions}.
All extracted annotations are manually inspected against the original papers to verify correctness and consistency.

{
\begin{researcherpromptbox}{\researcherbench{} - Instructions for Annotators}
We are building a new benchmark to evaluate the scientific research capabilities of language models (LLMs). Our current focus is on asking LLM agents to plan the ablation study of a paper, given its method section.

Your task is to identify and record ablation studies from a research paper.

Please follow the exact instructions to make our ground-truth exact and valid.

\textit{Please do NOT use any GenAI for the task.}

\smallskip
\textbf{What to include:}

\begin{itemize}
    \item Extract only experiments that are explicitly described by the authors as ablation studies.
    \item The paper must use the term ``ablation'' to describe the experiment.
    \item Ablation studies may appear anywhere in the paper, including the appendix.
\end{itemize}

\smallskip
\textbf{What to exclude:}

\begin{itemize}
    \item Do not include experiments that are not labeled as ablations by the authors, even if they look similar.
    \item Specifically exclude:
    \subitem Hyperparameter tuning experiments.
    \subitem Baseline or model comparisons.
    \subitem Any experiment not explicitly called an ablation.
\end{itemize}

\smallskip
\textbf{How to record each ablation:}

For each ablation study, create a JSON entry with the following fields:
\begin{enumerate}
    \item \texttt{name}: an indicative name provided by the authors.
    \item \texttt{ablated\_part}:  the component being ablated.
    \item \texttt{action}: one of \texttt{REMOVE}, \texttt{REPLACE}, or \texttt{ADD}, indicating the modification applied to the component.
    \item \texttt{replacement}: a list of components used as substitutes when the action is \texttt{REPLACE} or \texttt{ADD}.
    \item \texttt{metrics}: a list of evaluation metrics reported for the ablation experiment.
\end{enumerate}

\end{researcherpromptbox}
\captionsetup{type=figure}
\captionof{figure}{The instructions for the annotators of \researcherbench{}.} \label{fig:annotator_instructions}
}

Through the process of benchmark curation, the original 203 papers are filtered down to the 83 papers which comprise \researcherbench{}. We further split the 83 instances into two splits using random selection -- development split containing 21 papers and test split containing 62 papers.

\subsection{Constructing \reviewerbench{}} \label{app:reviewer_bench_construction} 
Open review policies adopted by conferences such as ICLR provide data to study the peer review process at scale, including reviewer-initiated suggestions for additional ablations. To construct the \reviewerbench{} benchmark, we use submissions to the ICLR conference from 2023 to 2025, focusing on cases where reviewers proposed ablation experiments that were not included in the original submission. This setup supports the development of reviewer-assistive models that can automatically identify potential gaps in a paper's empirical evaluation.

We curate the benchmark using three-stages pipeline:

\paragraph{1. Submission collection and basic filtering.} We collect all $\sim 89,100$ reviews associated with $\sim 22,800$ ICLR submissions from 2023 to 2025 and filter for those containing the substring "ablat".
    
\paragraph{2. Filtering for reviews suggesting new ablations.}
We apply an LM (GPT-4o) with CoT prompting (\Cref{fig:iclr_review_filter_prompt}) to identify reviews that explicitly suggest new ablation experiments not already included in the paper. We calibrate this filter on the development set by manually annotating the number of missing ablation suggestions appearing in each review and comparing them to the model predictions. On this set, the CoT-based filter achieves 88\% accuracy in predicting the number of suggested ablations, with a mean absolute error of 0.46.

\paragraph{3. Paper collection.} For each selected submission, we retrieve the submitted preprint PDF and convert it to markdown format using the \texttt{marker} tool~\cite{markertool}. This produces a standardized and machine-readable version of the full paper used as input to models.

{
\begin{reviewerpromptbox}{Instance Prompt}
You are given a peer review of a machine learning paper submitted to ICLR.
Your task is to determine how many new ablation study experiments the reviewer is suggesting that the paper is missing.

\smallskip
DEFINITION: 

A missing ablation refers to an experiment that the reviewer believes should have been conducted, in which a specific component, module, feature, or design choice is removed, replaced, or altered in order to assess its impact on the model's performance.
The review must clearly indicate that such an ablation study is missing or should be added.

\smallskip
You should answer with a number > 0 only if:

1. The review explicitly mentions the word ablation or ablate, and

2. It refers to a study that the paper did not perform, and

3. The reviewer is asking for that study to be added or noting that it is missing.

\smallskip
You should answer with a number = 0 if:

1. The review only discusses existing ablations.

2. The review mentions comparisons to baselines, datasets, or methods but not as ablation studies.

3. The word "ablation" appears in a different context (e.g., praising existing ablations).

4. The reviewer mentions the lack of ablation study but doesn't give any specific suggestions.

\smallskip
EXAMPLES:

1. "Lack of Ablation Study:  An ablation study investigating the impact of different components of the advanced memory module (e.g., summary length, specific aspects included) would provide a deeper understanding of its effectiveness. This would be an interesting result to many. AFAIK, lots and lots of memory mechanisms have been tested in many different contexts, but very few have offered insights on how memory mechanisms should be constructed.
." → The reviewer specifically mentions that the paper lacks an ablation study regarding the impact of different components of the advanced memory module, and mentions two possible components of it that should be ablated.  → 2

2. "Lack of ablation study." → The reviewer mentions the lack of ablation studies but doesn't suggest any specific ones. → 0

3. "The improvement over GeoDiff is not significant to me. Could the author provide more ablation study about the $f_\sigma$ function in Eq7, which can help to verify the importance of the proposed MB diffusion distribution." → The reviewer specifically mentions that the paper lacks an ablation study regarding the $f_\sigma$ function in Eq7, and mentions that it is important to verify the importance of the proposed MB diffusion distribution. → 1

4. "The experiments are limited. A more comprehensive evaluation is needed to trust the effectiveness of the proposed method." → The reviewer mentions the lack of experiments but doesn't refer to ablation study. → 0
\end{reviewerpromptbox}
\captionsetup{type=figure}
\captionof{figure}{The instance prompt for filtering ICLR reviews to get the number of ablation suggestion per each review (if any) for \reviewerbench{}{}.} \label{fig:iclr_review_filter_prompt}
}

Through the process of benchmark curation, the original set of 22,800 ICLR submissions was filtered down to the 5,960 papers included in \reviewerbench{}-\benchmark{}. Among these, 1,844 submissions were accepted (31\%) and 4,116 were rejected (69\%). We randomly select from the 5,960 papers a development set of 50 papers, and a test set of 300 papers.

\subsection{Licenses}

\researcherbench{} consists of papers from datasets with permissive licenses. CSR-Bench~\cite{xiao2025csrbenchbenchmarkingllmagents} includes only papers with permissive licenses as stated in their paper, SUPER-Expert~\cite{bogin2024superevaluatingagentssetting} is released under the Apache-2.0 license, and PaperBench~\cite{starace2025paperbenchevaluatingaisability} uses the MIT license. Additional papers from CVPR and ICCV are licensed under CC-BY-4.0. For the creation of \researcherbench{}, papers were modified through truncation, as permitted by their licenses. Papers were not anonymized, since authors can often be identified from the paper title and content.

For \reviewerbench{}, all selected papers and reviews are sourced from OpenReview and released under the CC-BY-4.0 license, in accordance with OpenReview's terms and conditions. Papers were collected in their anonymized versions whenever available, while reviews are always anonymized under ICLR's double-blind review policy. Papers and reviews were used as-is without modification.

\section{Properties of \benchmark{}}
\label{app:ablationbench_props}

\begin{table}[tb]
    
    \centering
    \resizebox{0.9\linewidth}{!}{
    \small
    \begin{tabular}{lrr}
    \toprule
        & Mean & Median \\
    \midrule
        (a) \# ablations in GT & &  \\
        \midtab \researcherbench{} & 3.7 & 3 \\
        \midtab \reviewerbench{} & 1.8 & 1\\
        \midrule
        (b) \# \TeX{} files in truncated paper & 4.2 & 2 \\
        \midrule
        (c) \# figures & & \\
        \midtab \researcherbench{} & 6.8 & 4 \\
        \midtab \reviewerbench{} & 9.4 & 8 \\
    \bottomrule
    \end{tabular}
    }
    \caption{
    Mean and median of attributes for a task instance in \benchmark{} test set. Row (b) is only relevant for \researcherbench{}.
    } \label{table:dataset_stats}
\end{table}

\begin{figure}[t]
\centering
\includegraphics[width=0.85\linewidth]{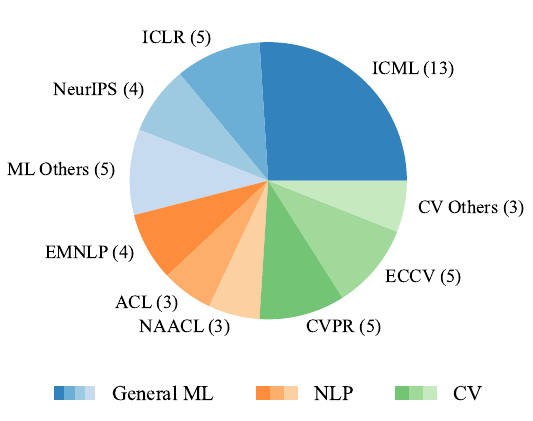}
\caption{Conference distribution of \researcherbench{} test set.}
\label{fig:conference_dist}
\end{figure}

\benchmark{}, composed of \researcherbench{} and \reviewerbench{}, offers a comprehensive evaluation suite for ablation planning in empirical AI research, covering diverse settings, topics, and ablation types. Key statistics of the test sets and benchmark characteristics are presented below.

\paragraph{Diversity of ablations.}
\Cref{table:dataset_stats}(a) shows the number of GT ablations in each benchmark. \researcherbench{} includes 230 ablations (4 per paper on average), with 59\% involving modifications and 41\% removals. Modification ablations often requiring multiple valid alternatives (3.3 on average). In \reviewerbench{}, papers contain on average 2 reviewer-suggested ablations, with over half containing only one.

\paragraph{Importance of ablation studies in review process.}
In \reviewerbench{}, nearly half (48\%) of all ICLR submissions between 2023 and 2025 include at least one review mentioning the keyword ``ablat'', with about one out of five reviews mention ablations explicitly. This indicates that ablations are a frequent point of feedback throughout the review process. Among the 300 submissions that comprise the test set of the benchmark, 31\% were accepted—matching the overall ICLR acceptance rate—suggesting that reviewer attention to ablations appears independent of final decisions.

\begin{figure}[t]
\centering
\includegraphics[width=0.9\linewidth]{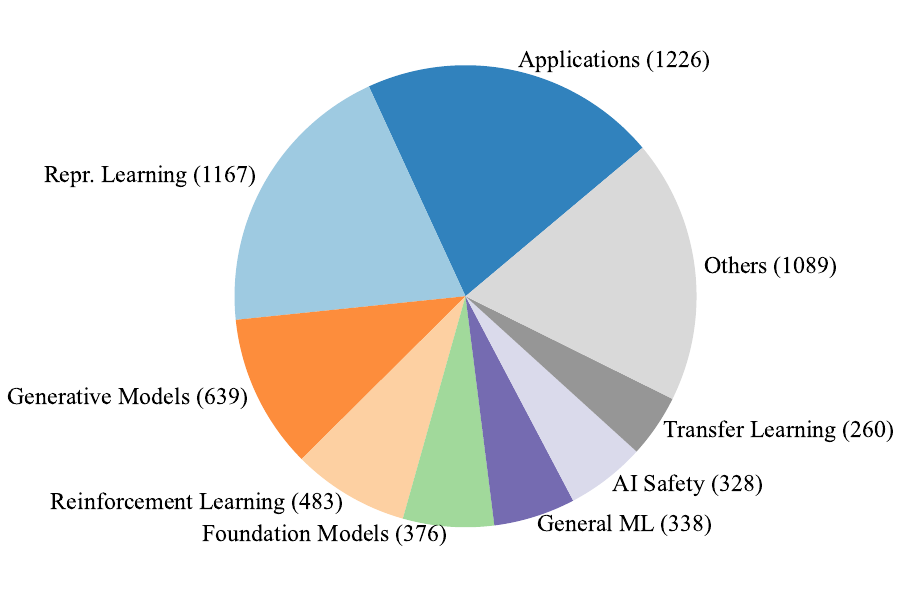}
\caption{Topic distribution of \reviewerbench{} full test set, extracted from OpenReview.}
\label{fig:topic_dist}
\end{figure}

\paragraph{Conferences and topics.}
\researcherbench{} spans across 14 conferences in multiple areas of AI research, including general machine learning conferences, NLP, and computer vision. The distribution of conferences is shown in~\Cref{fig:conference_dist}. \reviewerbench{} covers a diverse set of topics (\Cref{fig:topic_dist}) such as AI applications, representation learning, and generative models. This diversity across conferences and research areas enables \benchmark{} to support evaluation of ablation planning across a wide range of AI domains.

\newpage
\section{Experiments and Results} \label{app:ext_results}

This section presents additional results and analysis from the experiments in~\Cref{sec:experiments}, including per-task results, error analysis for both judges, and qualitative analysis of the planners. We observe that models that perform well on one task often perform poorly on the other: high-level reasoning is advantageous in \reviewerbench{}, allowing models like Llama to suggest potentially useful ablations even without deeply analyzing the paper, while fine-grained, detailed reasoning benefits \researcherbench{}, where precise analysis of method components is crucial. Overall, this illustrates how reasoning style influences planner performance across tasks.

\begin{table}[t]
    \centering
    \setlength{\tabcolsep}{2pt}
    \resizebox{0.9\linewidth}{!}{
    \begin{tabular}{lcccc}
    \toprule
        & Precision & Recall & F1 & $\kappa$ \\
    \midrule
        \textsc{LMJudge} & & & & \\
        \midtab w/ GPT-5.4 & 0.83 & 0.89 & 0.83 & 0.55 \\
        \midtab w/ o3-mini & 0.81 & 0.89 & 0.81 & 0.57 \\
        \midtab w/ GPT-OSS-120B & 0.81 & 0.88 & 0.79 & 0.43 \\
        \midtab w/ MiniMax-M2.5 & 0.79 & 0.86 & 0.76 & 0.47 \\
        \midtab w/ Qwen3.5-122B-A10B & 0.85 & 0.84 & 0.80 & 0.51 \\
        \midtab Majority Vote & \textbf{0.86} & \textbf{0.91} & \textbf{0.86} & \textbf{0.66} \\
        \cmidrule(ll){1-5}
        \textsc{AgentJudge} & & & & \\
        \midtab w/ GPT-5.4 & 0.79 & 0.88 & 0.79 & 0.43 \\
        \midtab w/ o3-mini & \textbf{0.86} & 0.85 & 0.80 & 0.51 \\
        \midtab w/ GPT-OSS-120B & 0.84 & \textbf{0.92} & \textbf{0.84} & 0.53 \\
        \midtab w/ MiniMax-M2.5 & 0.77 & 0.82 & 0.74 & 0.38 \\
        \midtab w/ Qwen3.5-122B-A10B & 0.81 & 0.91 & 0.80 & 0.42 \\
        \midtab Majority Vote & 0.85 & 0.88 & 0.82 & \textbf{0.54} \\
    \bottomrule
    \end{tabular}
    }
    \caption{\textbf{Evaluating \researcherbench{} Judge:} Macro-average results of the two judge scaffoldings measured on \researcherjudgedataset{}.}
    \label{table:researcher_judge_results}
\end{table}

\begin{table}[t]
    \centering
    \setlength{\tabcolsep}{2pt}
    \resizebox{0.9\linewidth}{!}{
    \begin{tabular}{lcccc}
    \toprule
        & Precision & Recall & F1 & $\kappa$ \\
    \midrule
        \textsc{LMJudge} & & & & \\
        \midtab w/ GPT-5.4 & 0.70 & 0.72 & 0.69 & 0.56 \\
        \midtab w/ o3-mini & 0.71 & \textbf{0.82} & \textbf{0.74} & 0.59 \\
        \midtab w/ GPT-OSS-120B & 0.58 & 0.62 & 0.47 & 0.47 \\
        \midtab w/ MiniMax-M2.5 & 0.69 & 0.74 & 0.69 & 0.54 \\
        \midtab w/ Qwen3.5-122B-A10B & \textbf{0.72} & 0.77 & 0.73 & \textbf{0.63} \\
        \midtab Majority Vote & 0.69 & 0.77 & 0.71 & 0.57 \\
        \cmidrule(ll){1-5}
        \textsc{AgentJudge} & & & & \\
        \midtab w/ GPT-5.4 & 0.60 & 0.59 & 0.58 & 0.46 \\
        \midtab w/ o3-mini & 0.69 & 0.71 & 0.68 & 0.55 \\
        \midtab w/ GPT-OSS-120B & 0.57 & 0.57 & 0.55 & 0.45 \\
        \midtab w/ MiniMax-M2.5 & \textbf{0.71} & \textbf{0.77} & \textbf{0.72} & 0.59 \\
        \midtab w/ Qwen3.5-122B-A10B & 0.64 & 0.63 & 0.63 & 0.53 \\
        \midtab Majority Vote & \textbf{0.71} & 0.70 & 0.69 & \textbf{0.59} \\
    \bottomrule
    \end{tabular}
    }
    \caption{\textbf{Evaluating \reviewerbench{} Judge:} Macro-average results of the two judge scaffoldings measured on \reviewerjudgedataset{}.}
    \label{table:reviewer_judge_results}
\end{table}

\begin{table*}[t]
    \centering
    \setlength{\tabcolsep}{2pt}
    
    \centering
    
    \vskip 0.15in
    \resizebox{0.7\textwidth}{!}{
    \small
    \begin{tabular}{lcccc}
    \toprule
        & Precision@5 $\uparrow$ & Recall@5 $\uparrow$ & F1@5 $\uparrow$ & nDCG@5 $\uparrow$ \\
    \midrule
        \textsc{LM-Planner} & & & & \\
        \midtab w/ GPT-5.4 & \textbf{0.35} & \textbf{0.52} & \textbf{0.39} & \textbf{0.53} \\
        \midtab w/ o3-mini & 0.31 & 0.42 & 0.33 & 0.45  \\
        \midtab w/ Gemini 2.5 Flash & 0.32 & 0.39 & 0.32 & 0.41 \\
        \midtab w/ Llama 3.1 405B Instruct & 0.23 & 0.33 & 0.25 & 0.34  \\
        \midtab w/ GPT-OSS-120B & 0.33 & 0.49 & 0.37 & 0.51  \\
        \midtab w/ MiniMax-M2.5 & 0.30 & 0.41 & 0.32 & 0.41 \\
        \midtab w/ Qwen3.5-122B-A10B & 0.33 & 0.48 & 0.36 &0.50  \\
        \cmidrule(ll){1-5}
        \textsc{Agent-Planner} & & & & \\
        \midtab w/ GPT-5.4 & \textbf{0.37} & \textbf{0.52} & \textbf{0.40} & \textbf{0.55} \\
        \midtab w/ Llama 3.1 405B Instruct & 0.23 & 0.22 & 0.21 & 0.23 \\
        \midtab w/ GPT-OSS-120B & 0.10 & 0.14 &0.11 &0.16  \\
        \midtab w/ MiniMax-M2.5 & 0.12 & 0.18 & 0.13 & 0.19  \\
        \midtab w/ Qwen3.5-122B-A10B & 0.33 & 0.45 & 0.35 & 0.47  \\
    \bottomrule
    \end{tabular}
    }
    \caption{\textbf{Evaluating \researcherbench{} Planners:} Macro-average results of the two planner scaffoldings.}
    \label{table:researcher_main_results}
\end{table*}

\begin{table}[t]
    \centering
    \setlength{\tabcolsep}{2pt}
    
    \centering
    
    \vskip 0.15in
    \resizebox{0.45\textwidth}{!}{
    \small
    \begin{tabular}{lccc}
    \toprule
        & Precision@2 $\uparrow$ & Recall@2 $\uparrow$ & F1@2 $\uparrow$ \\
    \midrule
        \textsc{LM-Planner} & & & \\
        \midtab w/ GPT-5.4 & \textbf{0.31} & \textbf{0.37} & \textbf{0.31}  \\
        \midtab w/ o3-mini & 0.20 & 0.24 & 0.20 \\
        \midtab w/ Gemini 2.5 Flash & 0.25 & 0.30 & 0.26 \\
         \midtab w/ Llama 3.1 405B Instruct & 0.25 & 0.32 & 0.27 \\
         \midtab w/ GPT-OSS-120B & 0.23 & 0.31 & 0.25  \\
        \midtab w/ MiniMax-M2.5 & 0.27 & 0.32 & 0.28  \\
        \midtab w/ Qwen3.5-122B-A10B & 0.23 & 0.29 & 0.24  \\
        \cmidrule(ll){1-4}
        \textsc{Agent-Planner} & & & \\
        \midtab w/ GPT-5.4 & \textbf{0.29} & \textbf{0.35} & \textbf{0.30}  \\
        \midtab w/ Llama 3.1 405B Instruct & 0.24 & 0.29 & 0.24 \\
        \midtab w/ GPT-OSS-120B & 0.21 & 0.25 & 0.22   \\
        \midtab w/ MiniMax-M2.5 & 0.24 & 0.29 & 0.25 \\
        \midtab w/ Qwen3.5-122B-A10B & 0.25 & 0.29&  0.25  \\
    \bottomrule
    \end{tabular}
    }
    \caption{\textbf{Evaluating \reviewerbench{} Planners:} Macro-average results of the two planner scaffoldings.}
    \label{table:reviewer_main_results}
\end{table}

\begin{table*}[tb]
\centering
\small
\begin{tabular}{lp{6cm}rr}
\toprule
\textbf{Error Type} & \textbf{Description} & \textbf{o3-mini (\%)} & \textbf{GPT~OSS~120B (\%)} \\
\midrule
\textbf{False Negative} & The judge marked it as "not found" even though the ablation appears in the prediction. & 51 & 17 \\
\textbf{Replacement Mismatch} & The correct component is identified, but the suggested replacements are not aligned with the GT. & 9 & 20 \\
\textbf{Surface-Level Match} & The match is based on similar wording, but the actual meaning does not align. & 7 & 30 \\
\textbf{Partial Match} & The predicted ablation is either broader or narrower than the GT. & 16 & 26 \\
\textbf{Action Type Mismatch} & The action (e.g., remove vs. modify) in the predicted ablation differs from the GT. & 12 & 5 \\
\textbf{Other} & Miscellaneous errors that do not fall into the above categories (e.g., formatting errors). & 5 & 2 \\
\bottomrule
\end{tabular}
\caption{Analysis of prediction errors for \researcherbench{}  judge, using \textsc{LMJudge} powered by o3-mini and GPT~OSS, measured on \researcherjudgedataset{}. Percentage indicates the percentage of all errors.}
\label{table:error_researcher_judge_analysis}
\end{table*}

\begin{table*}[tb]

\centering
\small
\begin{tabular}{p{3cm} p{6cm} r r}
\toprule
\textbf{Error Type} & \textbf{Description} & \textbf{o3-mini (\%)} & \textbf{GPT~OSS~120B (\%)} \\
\midrule
\textbf{Misinterpreted due to review breadth} & The review includes an ablation suggestion, but the model draws an incorrect conclusion due to vague or broad review suggestions. & 25 & 38 \\
\textbf{Ablation not identified} & The model fails to identify the ablation suggestion in the review (although it appears). & 34 & 37 \\
\textbf{Misinterpreted due to contextual understanding} & The model identifies the ablation but misinterprets it in the context of the paper. & 17 & 5 \\
\textbf{Incorrectly flags a non-ablation experiment} & The model mistakenly classifies an unrelated experiment suggestion as a suggested ablation. & 19 & 18 \\
\textbf{Other} & Miscellaneous errors that do not fall into the above categories (e.g., formatting errors). & 5 & 2 \\
\bottomrule
\end{tabular}
\caption{Analysis of prediction errors for \reviewerbench{}  judge, using \textsc{LMJudge} powered by o3-mini and GPT~OSS, measured on \reviewerjudgedataset{}. Percentage indicates the percentage of all errors.}
\label{table:error_reviewer_judge_analysis}
\end{table*}

\subsection{\researcherbench{}}

\begin{figure*}[tb]
    \centering
    \includegraphics[width=0.8\textwidth]{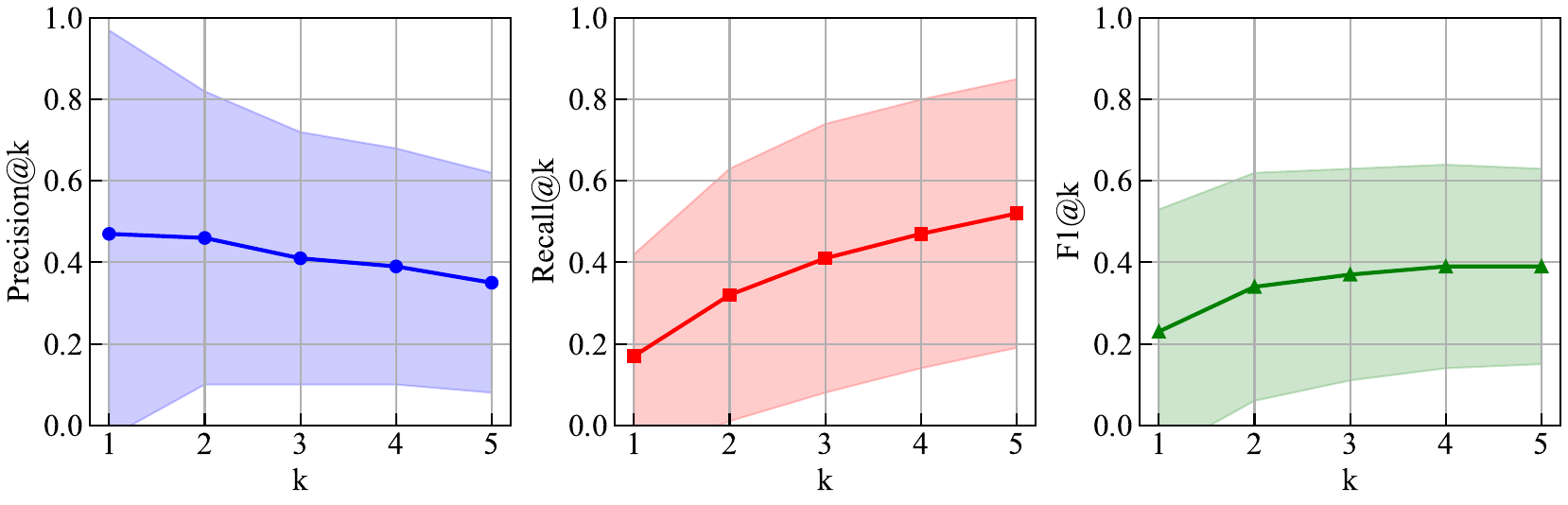}
    \caption{Evaluation metrics for \researcherbench{} using the \textsc{LM-Planner} with GPT-5.4, across varying values of $k$, number of generated ablations. Standard error presented across papers.}
    \label{fig:metrics_per_k}
\end{figure*}

\paragraph{Varying the number of generated ablations ($k$) has a minor effect on performance.}
\Cref{fig:metrics_per_k} shows the results for different values of $k$ using the best-performing model. As expected, precision decreases as $k$ increases, while recall improves. Overall, $k=4,5$ offers the best trade-off between precision and recall.

\paragraph{Models tend to recover important ablations first.} 
For \researcherbench{}, we additionally report \textit{nDCG@$k$}~\cite{10.1145/582415.582418}, which evaluates ranking quality in search algorithms. Since GT ablations are ordered by their appearance in the paper, we use nDCG@$k$ as a proxy for how well a model recovers more important ablations, under the assumption that the order of ablations in the paper roughly reflects their relative importance.
In our setting, we define relevance for nDCG@$k$ as a binary indicator of whether the ablation appears in the paper.
\Cref{table:researcher_main_results} indicates a strong correlation between nDCG@5 and Recall@5 (Pearson correlation of 0.97 on average across all models). This indicates that when models successfully recover ablations, they tend to prioritize those that appear earlier in the paper. Interpreting earlier appearance as a proxy for ablation importance, this suggests that models preferentially recover more central ablations before less prominent ones. 

\begin{figure*}[t]
    \centering
    \includegraphics[width=0.65\linewidth]{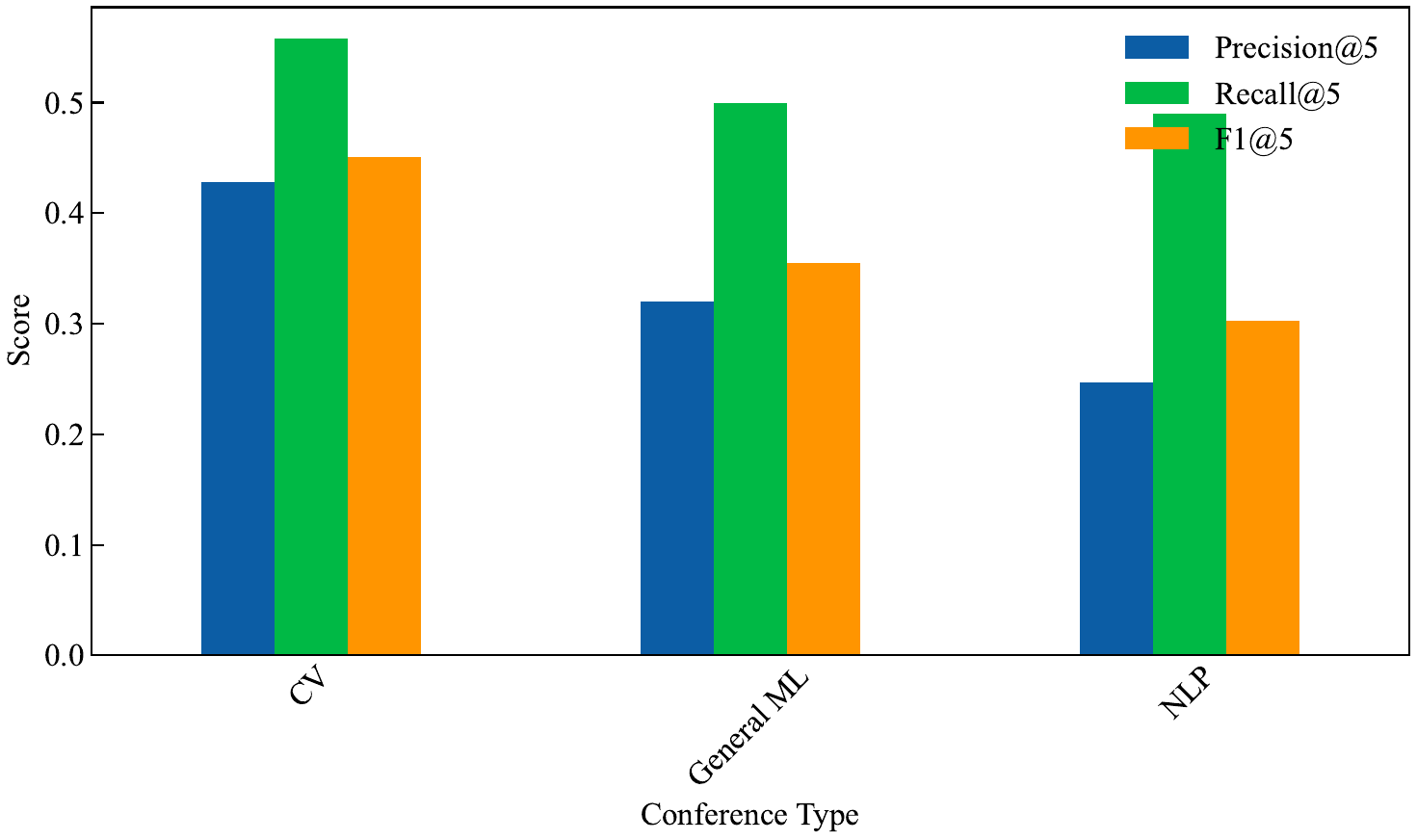}
    \caption{Metrics@5 per conference type measured on \researcherbench{} using the \textsc{LM-Planner} with GPT-5.4. We categorize the conferences in the following manner: (1) \textit{General ML:} ICML, NeurIPS, ICLR, AAAI, IJCAI, KDD; (2) \textit{CV:} CVPR, ECCV, ICPR, IEEE, ICCV; and (3) \textit{NLP:} ACL, EMNLP, NAACL.}
    \label{fig:metrics_per_conf_type}
\end{figure*}

\subsubsection{Qualitative Analysis} \label{sec:qualitative_researcher}
In the following, we qualitatively analyze the \textsc{LM-Planner} using Qwen~3.5 and Llama 3.1, on five randomly selected papers. This analysis examines the models' reasoning traces to understand how they plan ablations, which types of ablations they propose, and whether they suggest ablations that go beyond or improve upon those reported by the original authors. We provide reasoning traces for all papers as part of the supplementary material.

\paragraph{GPT-GNN: Generative Pre-Training of Graph Neural Networks~\cite{hu2020gptgnngenerativepretraininggraph}.} GPT-GNN proposes a framework to initialize GNNs by generative pre-training, by factorizing the likelihood of graph generation into two components: (1) attribute generation and (2) edge generation. The GT evaluation includes six ablations:
\begin{enumerate}
\item Removing the attribute generation component.
\item Removing the edge generation component.
\item Removing both attribute generation and node separation.
\item Removing edge generation together with the adaptive queue.
\item Replacing the GNN architecture (GCN, GAT, RGCN, HAN).
\item Varying the amount of labeled data during fine-tuning.
\end{enumerate}

Both models successfully identify the main components of the method: attribute generation, edge generation, node separation, and adaptive queue—which correspond to the core mechanisms targeted in the first four GT ablations.
Qwen's reasoning trace (~\Cref{fig:gpt_gnn_reasoning_Qwen}) shows a clear methodological analysis: it first identifies the main contributions and then proposes ablations based on this analysis. The model successfully recovers the first four GT ablations, all of these correspond to removal-type ablations, categories where models also perform best quantitatively (\Cref{fig:action_comparison}). It additionally suggests fine-grained modifications, such as changes to masking strategies or sampling procedures. While these ablations are relevant to the method, they are closer to hyperparameter tuning than true ablation studies, as they do not directly attribute performance to the paper's central contributions.
Llama, in contrast, produces a less structured reasoning trace (~\Cref{fig:gpt_gnn_reasoning_llama})  that mixes component identification with ablation suggestions. It favors removal-type ablations and recovers the main removal ablations for attribute generation, edge generation, and the adaptive queue. However, some of its proposals, such as removing the entire pre-training stage, do not isolate specific contributions of the method, providing limited insight into the mechanisms responsible for GPT-GNN's performance. Overall, Llama identifies the main components but designs ablations that are less precise and less aligned with the goals of a proper ablation study.

{
\begin{researcherpromptbox}{Reasoning - Qwen 3.5}
Based on the paper, GPT-GNN's main contributions are:

1. A novel attributed graph generation task that factorizes into Attribute Generation and Edge Generation

2. An efficient framework to conduct generative pre-training

3. Handling of large-scale graphs using sub-graph sampling and adaptive embedding queue

Key ablation studies should focus on:

1. The importance of the dual-generation approach (attributes and edges)

2. The effectiveness of the adaptive queue for negative samples

3. The impact of separating nodes into Attribute Generation and Edge Generation types

4. The sampling strategy for observed vs. masked edges

5. The sub-graph sampling approach for large graphs

I'll rank these by importance and formulate specific ablation experiments:
\end{researcherpromptbox}
\captionsetup{type=figure}
\captionof{figure}{The reasoning of \textsc{LM-Planner} with Qwen 3.5 in generating the ablation plans for GPT-GNN.} \label{fig:gpt_gnn_reasoning_Qwen}
}

{
\begin{researcherpromptbox}{Reasoning - Llama 3.1 405B}
To design ablation studies for the GPT-GNN framework, we need to consider the key components of the method and how they contribute to its overall performance. The GPT-GNN framework is designed for generative pre-training of graph neural networks, and it involves several important components, including the use of a self-supervised attributed graph generation task, the factorization of the graph generation objective into two components (Attribute Generation and Edge Generation), and the use of an adaptive node embedding queue to mitigate the inaccurate loss brought by negative sampling.

Based on these components, we can design several ablation studies to investigate their impact on the performance of the GPT-GNN framework.

1. **Ablation of the self-supervised attributed graph generation task**: In this ablation study, we can remove the self-supervised attributed graph generation task and instead use a traditional supervised learning approach to train the GNN model. This will allow us to investigate the importance of the self-supervised task in the GPT-GNN framework.

2. **Ablation of the Attribute Generation component**: In this ablation study, we can remove the Attribute Generation component and only use the Edge Generation component to generate the graph. This will allow us to investigate the importance of modeling the attributes of the nodes in the graph.

3. **Ablation of the Edge Generation component**: In this ablation study, we can remove the Edge Generation component and only use the Attribute Generation component to generate the graph. This will allow us to investigate the importance of modeling the edges between the nodes in the graph.

4. **Ablation of the adaptive node embedding queue**: In this ablation study, we can remove the adaptive node embedding queue and instead use a traditional negative sampling approach. This will allow us to investigate the importance of the adaptive node embedding queue in mitigating the inaccurate loss brought by negative sampling.

5. **Ablation of the pre-training task**: In this ablation study, we can remove the pre-training task and instead train the GNN model from scratch on the downstream task. This will allow us to investigate the importance of the pre-training task in the GPT-GNN framework.

By conducting these ablation studies, we can gain a better understanding of the importance of each component in the GPT-GNN framework and how they contribute to its overall performance.
\end{researcherpromptbox}
\captionsetup{type=figure}
\captionof{figure}{The reasoning of \textsc{LM-Planner} with Llama 3.1 405B in generating the ablation plans for GPT-GNN.} \label{fig:gpt_gnn_reasoning_llama}
}

\paragraph{SmoothNet: A Plug-and-Play Network for Refining Human Poses in Videos~\cite{zeng2022smoothnetplugandplaynetworkrefining}.} SmoothNet introduces a temporal-only refinement network designed to reduce jitter in video-based human pose estimation, for rarely seen or occluded actions. The paper evaluates its design through seven GT ablations that examine both architectural and training choices:
\begin{enumerate}
\item Replacing the FCN-based temporal model with alternative temporal models (Gaussian filtering, TCNs, and Transformers).
\item Replacing the proposed motion-aware temporal network with a simpler residual FCN.
\item Varying the temporal window size.
\item Removing the acceleration loss.
\item Removing the pose loss.
\item Replacing the motion modality representation.
\item Replacing the normalization strategy.
\end{enumerate}

Both models identify the key components of the method, including the motion-aware temporal network and temporal modeling choices. Qwen demonstrates strong alignment with the paper's main contributions (\Cref{fig:smoothnet_reasoning_Qwen}) and successfully recovers the majority of GT ablations, including the FCN-based temporal model, motion-aware network, window size, and loss-related ablations. It occasionally proposes deviations from the GT, such as joint-wise versus independent processing instead of the pose-loss removal; these ablations are still relevant but reflect a finer-grained focus on design choices rather than the high-level attribution of performance to main components.
Llama, by contrast, exhibits a less structured reasoning trace (\Cref{fig:smoothnet_reasoning_llama}), combining identification of components with ablation suggestions in a higher-level and imprecise manner. It confuses components and metrics, frames some ablations vaguely, and proposes ideas (e.g., transferability) that are not considered ablation studies. Consistent with the previous paper, Llama favors removal-type ablations and suggests broadly relevant research ideas rather than targeted experiments that isolate the contributions of specific components.

{
\begin{researcherpromptbox}{Reasoning - Qwen 3.5}
Based on the paper, SmoothNet is a temporal-only refinement network that aims to reduce jitters in human pose estimation. The key components that warrant investigation through ablation studies are:

1. The motion-aware architecture - The paper introduces velocity and acceleration branches which should be validated.

2. The temporal modeling approach - The authors chose FCN over TCN and Transformer, which needs validation.

3. The loss function components - The paper uses both pose and acceleration losses.

4. The temporal window size - This is a critical hyperparameter that affects the model's ability to capture long-range dependencies.

5. The independence of joint processing - The model processes each axis independently.

I'll suggest ablation studies to validate these key design choices, focusing on the most important aspects first.

The metrics used in the paper are:

- MPJPE (Mean Per Joint Position Error) for accuracy

- Accel (Mean Per Joint Acceleration Error) for smoothness
\end{researcherpromptbox}
\captionsetup{type=figure}
\captionof{figure}{The reasoning of \textsc{LM-Planner} with Qwen 3.5 in generating the ablation plans for SmoothNet.} \label{fig:smoothnet_reasoning_Qwen}
}

{
\begin{researcherpromptbox}{Reasoning - Llama 3.1 405B}
To design ablation studies for the SmoothNet paper, we need to identify the key components of the method and analyze their impact on the overall performance. The SmoothNet is a plug-and-play network for refining human poses in videos, which uses a temporal-only refinement approach to mitigate jitters. 

One potential ablation study is to analyze the effect of the motion-aware fully-connected network on the overall performance. This can be done by comparing the results of the basic SmoothNet with the motion-aware SmoothNet. 

Another potential ablation study is to investigate the impact of the long-range temporal receptive fields on the performance of the SmoothNet. This can be done by varying the length of the input sequence and analyzing the effect on the results.

Additionally, we can also analyze the effect of the loss function on the performance of the SmoothNet. This can be done by comparing the results of the SmoothNet trained with different loss functions, such as the mean per joint position error (MPJPE) and the mean per joint acceleration error (Accel).

Finally, we can also investigate the transferability of the SmoothNet across different datasets and estimators. This can be done by training the SmoothNet on one dataset and testing it on another dataset, or by using different estimators as input to the SmoothNet.
\end{researcherpromptbox}
\captionsetup{type=figure}
\captionof{figure}{The reasoning of \textsc{LM-Planner} with Llama 3.1 405B in generating the ablation plans for SmoothNet.} \label{fig:smoothnet_reasoning_llama}
}

\paragraph{Image Processing GNN: Breaking Rigidity in Super-Resolution~\cite{Tian_2024_CVPR}.}
Image Processing GNN (IPG) proposes a graph-based model for image super-resolution reconstruction that aims to break rigidity that dominates previous super-resolution methods, by leveraging degree flexibility in graphs, pixel node sets instead of patch nodes and local and global information. The paper evaluates its claims using four ground-truth ablations:
\begin{enumerate}
\item Replacing pixel-level nodes with patch-based nodes.
\item Replacing degree-flexible graphs with KNN graphs.
\item Removing the local graph construction.
\item Removing the global graph construction.
\end{enumerate}

Qwen demonstrates strong alignment with the paper's method (\Cref{fig:ipg_reasoning_Qwen}), explicitly identifying all three key components and successfully recovering all four GT ablations. Since the planner is prompted to produce up to five ablations, Qwen also proposes an additional ablation on auxiliary components, such as the detail-rich indicator and relative position encoding. While these go beyond the GT set, they remain relevant to the method's internal mechanisms and coherent with the research. Overall, Qwen generates a well-scoped ablation plan that directly attributes performance to the primary components responsible for breaking rigidity.
Llama, in contrast, exhibits a higher-level and less detailed reasoning trace (\Cref{fig:ipg_reasoning_llama}). It correctly identifies the main components and proposes ablations broadly aligned with the GT, particularly for degree flexibility, pixel versus patch nodes, and local versus global graphs. However, the ablations are described less concretely, and the model favors removal-type or conceptual ablations over precise experimental designs. As in prior cases, Llama occasionally suggests additional relevant modifications, such as graph aggregation and positional encoding, which are coherent with the research.

{
\begin{researcherpromptbox}{Reasoning - Qwen 3.5}
The IPG model introduces several novel components that contribute to "breaking rigidity" in super-resolution. The main innovations are:

1. Degree Flexibility: Using a detail-rich indicator (D\_F) to assign different node degrees to pixels based on their reconstruction needs.

2. Pixel-level Nodes: Using pixels as graph nodes instead of patches to avoid misalignment issues.

3. Dual-scale Graph Construction: Using both local and global sampling strategies to construct graphs at different scales.

Based on these key components, I suggest the following ablation studies:

1. Test the importance of degree flexibility by comparing against fixed degree for all nodes

2. Validate the benefit of pixel-level nodes vs patch-level nodes

3. Evaluate the contribution of dual-scale graphs vs single-scale

4. Examine the effectiveness of the detail-rich indicator metric

5. Test the importance of relative position encoding

I'll prioritize the most critical ones that directly validate the paper's main claims about breaking rigidity.
\end{researcherpromptbox}
\captionsetup{type=figure}
\captionof{figure}{The reasoning of \textsc{LM-Planner} with Qwen 3.5 in generating the ablation plans for IPG.} \label{fig:ipg_reasoning_Qwen}
}

{
\begin{researcherpromptbox}{Reasoning - Llama 3.1 405B}
The paper proposes a novel method for image super-resolution using a graph neural network (GNN) approach, called Image Processing GNN (IPG). The IPG model leverages the flexibility of graphs to break the rigidity of traditional super-resolution methods. The authors identify three types of rigidity in existing methods: degree-rigidity, patch-rigidity, and space-rigidity. To address these limitations, the IPG model uses a degree-flexible graph solution, pixel nodes instead of patches, and local and global sampling strategies.

To further investigate the effectiveness of the IPG model, I suggest the following ablation studies:

1. **Degree flexibility**: Remove the degree flexibility component and use a fixed degree for all nodes. This will help to understand the impact of degree flexibility on the model's performance.

2. **Pixel nodes vs. patch nodes**: Replace pixel nodes with patch nodes to see how it affects the model's performance. This will help to understand the importance of using pixel nodes in the IPG model.

3. **Local and global sampling**: Remove the local sampling strategy and only use global sampling. This will help to understand the impact of local sampling on the model's performance.

4. **Graph aggregation**: Replace the edge-conditioned aggregation with a different aggregation method, such as max-pooling. This will help to understand the impact of the aggregation method on the model's performance.

5. **Relative position encoding**: Remove the relative position encoding component and see how it affects the model's performance. This will help to understand the importance of relative position encoding in the IPG model.

These ablation studies will help to understand the contribution of each component to the IPG model's performance and provide insights for future improvements.

\end{researcherpromptbox}
\captionsetup{type=figure}
\captionof{figure}{The reasoning of \textsc{LM-Planner} with Llama 3.1 405B in generating the ablation plans for IPG.} \label{fig:ipg_reasoning_llama}
}

\paragraph{Delving into Deep Imbalanced Regression~\cite{yang2021delvingdeepimbalancedregression}.}
This paper proposes two complementary techniques for imbalanced regression: Label Distribution Smoothing (LDS) and Feature Distribution Smoothing (FDS), which exploit the similarity
between nearby targets in both label and feature space. The paper includes four GT ablation studies that focus on design choices within these components:
\begin{enumerate}
\item Replacing the kernel type used in LDS and FDS (e.g., Gaussian, Laplacian, Triangular).
\item Replacing the regression loss function used jointly with LDS and FDS.
\item Varying kernel hyperparameters (kernel size and standard deviation).
\item Testing robustness under different skewed label distributions.
\end{enumerate}

Both models correctly identify LDS and FDS as the core contributions of the paper, and they recover the kernel-type replacement ablation present in the GT. This shows that both models can capture the primary components and the corresponding ablation space. Beyond this, the models diverge in their approach: Qwen (~\Cref{fig:imbalance_regression_reasoning_Qwen}) proposes additional high-level removal or combination ablations, isolating LDS or FDS, which are relevant for analyzing component importance but do not appear in the GT. Llama (~\Cref{fig:imbalanced_regression_reasoning_llama}), in contrast, produces a less structured reasoning trace and frequently combines component identification with ablation suggestions. Its proposed ablations include removal-type studies, baseline comparisons, and alternative calibration methods that fall outside the intended scope of ablation studies, reflecting broader evaluation ideas rather than component-level analysis.

{
\begin{researcherpromptbox}{Reasoning - Qwen 3.5}
The paper introduces two main components to handle imbalanced regression:

1. Label Distribution Smoothing (LDS) - which uses kernel density estimation to learn the effective imbalance in datasets with continuous targets

2. Feature Distribution Smoothing (FDS) - which performs distribution smoothing on the feature space to calibrate potentially biased feature estimates

Based on these components, I'll suggest ablation studies focusing on:

1. The impact of kernel choice in LDS since it's fundamental to how the label distribution is smoothed

2. The effectiveness of FDS's momentum update mechanism

3. The necessity of feature calibration in FDS

4. The importance of bin size selection for grouping features

5. The combined effect of LDS and FDS versus using them individually

These ablations will help understand which components contribute most to the method's performance and how sensitive it is to different design choices.
\end{researcherpromptbox}
\captionsetup{type=figure}
\captionof{figure}{The reasoning of \textsc{LM-Planner} with Qwen 3.5 in generating the ablation plans for Delving into Deep Imbalanced Regression.} \label{fig:imbalance_regression_reasoning_Qwen}
}

{
\begin{researcherpromptbox}{Reasoning - Llama 3.1 405B}
The paper "Delving into Deep Imbalanced Regression" presents a novel approach to handling imbalanced data in regression tasks. The authors propose two techniques: Label Distribution Smoothing (LDS) and Feature Distribution Smoothing (FDS). LDS involves convolving a symmetric kernel with the empirical label density to estimate the effective label density distribution, while FDS performs distribution smoothing on the feature space to calibrate the potentially biased estimates of feature distribution.

To further investigate the effectiveness of these techniques, I suggest the following ablation studies:

1. **Remove LDS**: Train the model without LDS to see how much it contributes to the overall performance. This will help understand the impact of LDS on the model's ability to handle imbalanced data.

2. **Replace FDS with a different feature calibration method**: Replace FDS with a different feature calibration method, such as batch normalization or instance normalization, to see if FDS is the most effective method for feature calibration.

3. **Add a different kernel to LDS**: Add a different kernel, such as a Gaussian or Laplacian kernel, to LDS to see if the choice of kernel affects the performance of the model.

4. **Apply FDS to a different layer**: Apply FDS to a different layer of the network, such as the input layer or a hidden layer, to see if the location of FDS affects the performance of the model.

5. **Compare with other imbalanced regression methods**: Compare the performance of the proposed method with other imbalanced regression methods, such as SMOTE or cost-sensitive learning, to see if the proposed method is more effective.

These ablation studies will help understand the contributions of each component of the proposed method and provide insights into how to further improve its performance.
\end{researcherpromptbox}
\captionsetup{type=figure}
\captionof{figure}{The reasoning of \textsc{LM-Planner} with Llama 3.1 405B in generating the ablation plans for Delving into Deep Imbalanced Regression.} \label{fig:imbalanced_regression_reasoning_llama}
}

\paragraph{Neural-Pull: Learning Signed Distance Functions from Point Clouds by Learning to Pull Space onto Surfaces~\cite{ma2021neuralpulllearningsigneddistance}.}
Neural-Pull proposes learning signed distance functions directly from point clouds by iteratively pulling query locations toward the surface using predicted SDF values and gradients. The paper evaluates the method using three GT ablations that target core design choices:
\begin{enumerate}
\item Removing the geometric network initialization (GNI).
\item Replacing the query location sampling strategy with random sampling over the full 3D space.
\item Adding a gradient constraint that enforces unit-norm gradients.
\end{enumerate}

Both models correctly identify the main components of Neural-Pull, including the pulling operation, query location sampling strategy, and geometric network initialization, and they recover several of the GT ablations, such as the GNI removal and space-sampling replacement. Beyond these shared observations, the models differ in their handling of gradient-related ablations: Qwen (~\Cref{fig:neuralpull_reasoning_Qwen}) fails to recover the gradient-constraint ablation exactly as in the GT, instead proposing alternative optimization-focused ablations such as modifying the loss function or gradient computation, which remain relevant and coherent. Llama (~\Cref{fig:neuralpull_reasoning_llama}) partially overlaps with the GT on gradient and sampling ablations but reframes the gradient-constraint ablation as removing gradient information entirely. As in previous cases, Llama's proposals are generally higher-level, less structured, and focus on removal-type ablations.

{
\begin{researcherpromptbox}{Reasoning - Qwen 3.5}
Based on the paper, Neural-Pull's main contribution is learning SDFs directly from point clouds by pulling query locations to their closest points on the surface. The key components are:

1. The pulling operation that uses both predicted signed distance values and gradients
2. The adaptive sampling strategy for query locations using isotropic Gaussian functions
3. The geometric network initialization (GNI) to approximate sphere SDFs
4. The loss function based on minimizing distance between pulled locations and nearest neighbors

I'll suggest ablations that investigate these key components:

1. Test the importance of the pulling operation by comparing with direct distance prediction
2. Evaluate different query location sampling strategies
3. Assess the impact of the geometric network initialization
4. Examine alternative loss formulations
5. Test different gradient computation approaches

I'll rank these by importance based on their potential impact on the method's performance.
\end{researcherpromptbox}
\captionsetup{type=figure}
\captionof{figure}{The reasoning of \textsc{LM-Planner} with Qwen 3.5 in generating the ablation plans for Neural-Pull.} \label{fig:neuralpull_reasoning_Qwen}
}

{
\begin{researcherpromptbox}{Reasoning - Llama 3.1 405B}
The paper "NeuralPull: Learning Signed Distance Functions from Point Clouds by Learning to Pull Space onto Surfaces" presents a novel approach to learning signed distance functions (SDFs) from raw point clouds without requiring ground truth signed distance values. The method, called Neural-Pull, trains a neural network to pull query 3D locations to their closest points on the surface using the predicted signed distance values and the gradient at the query locations. To investigate the effectiveness of Neural-Pull, we suggest the following ablation studies:

1. **Removing the gradient component**: We will remove the gradient component from the pulling operation and only use the predicted signed distance values to move the query locations. This will help us understand the importance of the gradient information in the Neural-Pull method.

2. **Replacing the pulling operation with a different transformation**: We will replace the pulling operation with a different transformation, such as a simple translation or scaling, to see if the Neural-Pull method is sensitive to the choice of transformation.

3. **Adding noise to the point cloud**: We will add noise to the point cloud and evaluate the performance of Neural-Pull. This will help us understand the robustness of the method to noisy data.

4. **Using a different loss function**: We will replace the mean squared error loss function used in Neural-Pull with a different loss function, such as mean absolute error or cross-entropy loss, to see if the choice of loss function affects the performance of the method.

5. **Evaluating the effect of query location sampling**: We will evaluate the effect of different query location sampling strategies on the performance of Neural-Pull. This will help us understand the importance of the sampling strategy in the method.
\end{researcherpromptbox}
\captionsetup{type=figure}
\captionof{figure}{The reasoning of \textsc{LM-Planner} with Llama 3.1 405B in generating the ablation plans for Neural-Pull.} \label{fig:neuralpull_reasoning_llama}
}

\paragraph{Preserving Commonsense Knowledge from Pre-trained Language Models via Causal Inference~\cite{zheng2023preservingcommonsenseknowledgepretrained}.}
The paper proposes Causal Effect Tuning (CET) to preserve pre-trained models knowledge during fine-tuning and avoid overfitting. The GT includes a single ablation study focused on the KNN-based sample selection and search method, replaced with alternative selection strategies and similarity metrics, evaluated on average accuracy over commonsense QA datasets.

Both models correctly identify the main components of the method, including the KNN module, the causal framework, similarity threshold, and weighting scheme. They also recover aspects of the GT ablation space to some extent, particularly the KNN-based interventions. Beyond this shared observation, Qwen (~\Cref{fig:cet_reasoning_Qwen}) consistently proposes five ablations, even when the space of meaningful GT ablations is smaller, and tends to focus on fine-grained parameter tuning (e.g., the number of neighbors or the similarity threshold). These ablations are coherent and informative but do not fully target the methodologically critical component of replacing KNN for causal effect estimation. Llama (\Cref{fig:cet_reasoning_llama}) exhibits similar patterns as in previous papers, combining component identification with ablation proposals, with its suggestions being more naive (e.g., uniform estimation replacing KNN) and generally emphasizing hyperparameter-type changes rather than ablations that test the method's core functionality. As in previous papers, Llama produces ideas that are relevant to the field but not strictly aligned with the GT or classical ablation study goals.

{
\begin{researcherpromptbox}{Reasoning - Qwen 3.5}
Based on the paper, the main components of the method are:

1. The causal graph framework for fine-tuning that introduces colliding effects between pre-trained data and target data.

2. The K-Nearest Neighbors (KNN) approach to estimate causal effects, where neighbors are selected based on similarity of gold answers.

3. The unified objective that combines vanilla fine-tuning with causal effects preservation.

4. The similarity threshold $\Theta$ that controls which samples get KNN treatment.

5. The weighting parameter $W_0$ that balances between anchor sample and its neighbors.

I'll suggest ablation studies focusing on these key components, ranked by importance:

1. First, we should verify the effectiveness of the causal framework by removing it completely.

2. The KNN approach and similarity threshold are critical design choices that should be tested.

3. The weighting scheme between anchor and neighbors is important for balancing knowledge preservation.

4. Alternative similarity metrics for finding neighbors could be explored.

The metrics used in the paper are primarily accuracy on commonsense QA datasets, so we'll use those.
\end{researcherpromptbox}
\captionsetup{type=figure}
\captionof{figure}{The reasoning of \textsc{LM-Planner} with Qwen 3.5 in generating the ablation plans for Preserving Commonsense Knowledge from Pre-trained Language Models via Causal Inference.} \label{fig:cet_reasoning_Qwen}
}

{
\begin{researcherpromptbox}{Reasoning - Llama 3.1 405B}
To design ablation studies for the research titled "Preserving Commonsense Knowledge from Pre-trained Language Models via Causal Inference," we need to identify the key components of the proposed method, Causal Effect Tuning (CET), and analyze their impact on the model's performance. Here are five ablation study suggestions, ranked by their importance:

1. **Ablation of the causal objective**: Remove the causal objective from the overall objective function and fine-tune the model using only the vanilla fine-tuning objective. This ablation study will help us understand the contribution of the causal objective in preserving commonsense knowledge.

2. **Ablation of the KNN-based colliding effect estimation**: Replace the KNN-based colliding effect estimation with a random or uniform estimation method. This ablation study will help us understand the importance of using KNNs to estimate the colliding effect.

3. **Ablation of the similarity threshold**: Remove the similarity threshold ($\Theta$) and use a fixed number of nearest neighbors for all samples. This ablation study will help us understand the impact of the similarity threshold on the model's performance.

4. **Ablation of the weight controlling the strength of colliding effects (W0)**: Set W0 to 0 or 1 and analyze the impact on the model's performance. This ablation study will help us understand the importance of controlling the strength of colliding effects.

5. **Ablation of the heuristic approximation for estimating the causal effect on commonsense QA**: Replace the heuristic approximation with a different method, such as using the euclidean distance or cosine similarity between the average last hidden states of PLMs. This ablation study will help us understand the effectiveness of the proposed heuristic approximation.
\end{researcherpromptbox}
\captionsetup{type=figure}
\captionof{figure}{The reasoning of \textsc{LM-Planner} with Llama 3.1 405B in generating the ablation plans for Preserving Commonsense Knowledge from Pre-trained Language Models via Causal Inference.} \label{fig:cet_reasoning_llama}
}

\subsection{\reviewerbench{}}

\begin{table}[t]
    \centering
    \setlength{\tabcolsep}{2pt}
    \begin{minipage}[t]{0.48\textwidth}
        \centering

        \resizebox{0.8\textwidth}{!}{
        \small
        \begin{tabular}{lcc}
        \toprule
            Year & 2023-2024 & 2025 \\
        \midrule
            \textsc{LM-Planner} & & \\
            \midtab w/ GPT-5.4 & 0.33 &  0.39 \\
            \midtab w/ Llama 3.1 405B Instruct & 0.35 & 0.30  \\
            \midtab w/ o3-mini & 0.21 & 0.26  \\
            \midtab w/ Gemini 2.5 Flash & 0.30 & 0.30 \\
            \midtab w/ GPT~OSS~120B & 0.30 & 0.32 \\
            \midtab w/ MiniMax~M2.5 & 0.33 & 0.32 \\
            \midtab w/ Qwen 3.5 122B & 0.27 & 0.31 \\
            \cmidrule(ll){1-3}
            \textsc{Agent-Planner} & &\\
            \midtab w/ GPT-5.4 & 0.38 & 0.33 \\
            \midtab w/ Llama 3.1 405B Instruct & 0.29 & 0.29 \\
            \midtab w/ GPT~OSS~120B & 0.25 & 0.25 \\
            \midtab w/ MiniMax~M2.5 & 0.29 & 0.29 \\
            \midtab w/ Qwen 3.5 122B & 0.30 & 0.29 \\
        \bottomrule
        \end{tabular}
        }
        \captionof{table}{Macro-average recall@2 results of the two planner scaffoldings for the \reviewerbench{} split by year.}
        \label{table:results_per_year}
    \end{minipage}
    \hfill
    \begin{minipage}[t]{0.48\textwidth}
        \centering

        \resizebox{0.8\textwidth}{!}{
        \small
        \begin{tabular}{lcc}
        \toprule
            Decision & Accept & Reject \\
        \midrule
            \textsc{LM-Planner} & & \\
            \midtab w/ GPT-5.4 & 0.32 & 0.39 \\
            \midtab w/ Llama 3.1 405B Instruct & 0.31 & 0.33  \\
            \midtab w/ o3-mini & 0.25 & 0.23  \\
            \midtab w/ Gemini 2.5 Flash & 0.21 & 0.34 \\
            \midtab w/ GPT~OSS~120B & 0.26 & 0.33 \\
            \midtab w/ MiniMax~M2.5 & 0.31 & 0.33 \\
            \midtab w/ Qwen 3.5 122B & 0.28 & 0.30 \\
            \cmidrule(ll){1-3}
            \textsc{Agent-Planner} & &\\
            \midtab w/ GPT-5.4 & 0.27 & 0.39 \\
            \midtab w/ Llama 3.1 405B Instruct & 0.27 & 0.30 \\
            \midtab w/ GPT~OSS~120B & 0.21 & 0.27 \\
            \midtab w/ MiniMax~M2.5 & 0.21 & 0.32 \\
            \midtab w/ Qwen 3.5 122B & 0.28 & 0.30 \\
        \bottomrule
        \end{tabular}
        }
        \captionof{table}{Macro-average recall@2 results of the two planner scaffoldings for the \reviewerbench{} split by decision.}
        \label{table:results_per_decision}
    \end{minipage}
\end{table}

\paragraph{Models perform better on rejected papers.}
We test whether planner recall correlates with paper acceptance. Most models achieve higher recall on rejected papers (\Cref{table:results_per_decision}). This is expected: lower-quality papers often lack rigor and validation, making missing ablations easier to identify.

\paragraph{Results by year.}
\Cref{table:results_per_year} compares recall@2 on submissions from before 2025 and from 2025. Results are similar across both splits and show no consistent trend across models. This suggests that model performance is not strongly affected by the submission year, indicating generalization beyond the training data.

\paragraph{Results per acceptance status.}
\Cref{table:results_per_decision} reports recall@2 by acceptance status. Most models perform better on rejected papers than accepted ones. This suggests performance correlates with paper quality, and that proposing ablations is easier for lower-quality papers.

\subsubsection{Qualitative Analysis} \label{sec:qualitative_reviewer}

In the following, we qualitatively analyze the \textsc{LM-Planner} using Llama 3.1 and Qwen 3.5, on five randomly selected papers. This analysis examines the models' reasoning traces to understand how they find missing ablations, and which types of ablations they propose.

\paragraph{Boosting Membership Inference Attacks with Upstream Modification~\cite{abdullah2025boosting}.} The paper improves Membership Inference Attacks (MIAs) by modifying dataset partitioning and focusing attacks on outliers, achieving higher true positive rates and near-perfect AUC. No formal ablation studies are included. Llama correctly identifies this gap (\Cref{fig:boosting_reasoning_llama}) and proposes two high-level ablations: isolating the effect of reduced drop rate and evaluating the impact of using only outliers, effectively targeting the main method components, although not appearing in any review. Qwen, in contrast, produces a structured reasoning trace (\Cref{fig:boosting_reasoning_Qwen}) that first identifies the method's components and then incorrectly treats parameter sweeps (e.g., drop rates, number of shadow models, outlier detection settings) as ablations. Subsequently, it proposes more granular ablations focusing on hyperparameters, rather than the main modifications that determine the attack's effectiveness. Qwen's fine-grained analysis, while coherent, fails to capture the essence of ablation studies.

{
\begin{reviewerpromptbox}{Reasoning - Llama 3.1 405B}
The paper "Boosting Membership Inference Attacks with Upstream Modification" presents a novel approach to improve the accuracy of membership inference attacks (MIAs) on machine learning models. The authors identify two limitations in the existing MIA framework: sampling bias and attack aggregation. They propose modifications to address these limitations, which involve reducing the drop rate during dataset partitioning and running the MIA against only the outliers.

The authors demonstrate the effectiveness of their modifications through experiments on various datasets, including CIFAR-10, CIFAR-100, and Tiny-Imagenet. The results show that their modified attacks outperform the original ones in every case, achieving near-perfect AUC and TPR at zero FPR.

However, the paper lacks ablation studies to further investigate the impact of each modification on the attack success. Specifically, it would be interesting to see how the attack performance changes when only one of the modifications is applied.

To address this, I suggest two missing ablation studies:

1. **Ablation A:** Evaluate the impact of reducing the drop rate during dataset partitioning on the attack success, without modifying the attack aggregation step. This will help understand the contribution of the sampling bias modification to the overall improvement in attack accuracy.

2. **Ablation B:** Evaluate the impact of using only the outliers for the MIA, without modifying the sampling bias. This will help understand the contribution of the attack aggregation modification to the overall improvement in attack accuracy.

These ablation studies will provide a more detailed understanding of the effects of each modification and help identify potential areas for further improvement.
\end{reviewerpromptbox}
\captionsetup{type=figure}
\captionof{figure}{The reasoning of \textsc{LM-Planner} with Llama 3.1 405B in generating missing ablations for Boosting Membership Inference Attacks with Upstream Modification.}
\label{fig:boosting_reasoning_llama}
}

{
\begin{reviewerpromptbox}{Reasoning - Qwen 3.5}
The paper proposes modifications to the upstream stages of Membership Inference Attacks (MIAs), specifically focusing on two main components:

1. Sampling bias modification (Step 2): They reduce the drop rate during dataset partitioning from 50%

2. Attack aggregation modification (Step 4): They use outlier detection methods (specifically ApB - Accuracy per Batch) to identify vulnerable points instead of using the entire dataset.

The paper already includes ablations for:
- Different drop rates (50\%, 40\%, 30\%, 20\%, 10\%)

- Different outlier detection methods (SPP, MD, ApB)

- Different number of shadow models (1 vs 100)

However, there are two important missing ablations:

1. The effect of the number of vulnerable points selected: The paper fixes this to 5,000 points but doesn't ablate this hyperparameter. This is crucial since it directly affects both the computational cost and attack effectiveness.

2. The robustness of the ApB outlier detection method: While ApB performs best, the paper doesn't ablate its key parameter - the frequency of accuracy measurement (currently done after each batch). This is important since it affects both computational overhead and outlier detection accuracy.

These ablations would provide important insights into the method's sensitivity to these key parameters.
\end{reviewerpromptbox}
\captionsetup{type=figure}
\captionof{figure}{The reasoning of \textsc{LM-Planner} with Qwen 3.5 in generating missing ablations for Boosting Membership Inference Attacks with Upstream Modification.}
\label{fig:boosting_reasoning_Qwen}
}

\paragraph{Accelerating Retrieval-augmented Language Model Serving with Speculation~\cite{zhang2024accelerating}.} RaLMSpec accelerates retrieval-augmented language model serving using caching, batched verification, and speculation stride optimization. Both models miss an ablation that already appears in the paper (cache size variation). Llama's reasoning (\Cref{fig:accelerating_reasoning_llama}) describes the paper but ignores the existing ablations and proposes two experiments, one of which coincides with an existing ablation and the other aligns with a reviewer suggestion. Qwen provides a less structured reasoning (\Cref{fig:accelerating_reasoning_Qwen}), noting only that extensive experiments were performed. Its suggested ablations are sensible but either duplicate experiments already in the paper (e.g., cache size variations) or focus on parameter changes rather than altering components to yield insight into the method's behavior.

{
\begin{reviewerpromptbox}{Reasoning - Llama 3.1 405B}
The paper "Accelerating Retrieval-augmented Language Model Serving with Speculation" proposes a speculation-inspired framework called RaLMSpec to accelerate the serving of retrieval-augmented language models (RaLM). RaLMSpec uses speculative retrieval and batched verification to reduce the serving latency of iterative RaLM approaches while preserving the same model outputs. The framework also incorporates prefetching, optimal speculation stride scheduler, and asynchronous verification to further boost the speculation performance.

The paper presents extensive evaluations of RaLMSpec over three language models on four downstream QA datasets, demonstrating that RaLMSpec can achieve a speed-up ratio of 1.75-2.39×, 1.04-1.39×, and 1.31-1.77× when the retriever is an exact dense retriever, approximate dense retriever, and sparse retriever, respectively, compared with the baseline.

To further investigate the effectiveness of RaLMSpec, we suggest two missing ablation studies:

1. **Ablation study on the impact of speculation stride on the performance of RaLMSpec**: The paper mentions that the speculation stride is a crucial parameter that affects the trade-off between speculation overhead and latency saving. However, it does not provide a detailed analysis of how different speculation strides affect the performance of RaLMSpec. An ablation study on the impact of speculation stride can provide valuable insights into the optimal speculation stride for different scenarios.

2. **Ablation study on the effect of cache size on the performance of RaLMSpec**: The paper mentions that RaLMSpec uses a local cache to store retrieved documents, but it does not investigate the effect of cache size on the performance of RaLMSpec. An ablation study on the effect of cache size can help understand the optimal cache size for different scenarios and provide insights into the trade-off between cache size and performance.
\end{reviewerpromptbox}
\captionsetup{type=figure}
\captionof{figure}{The reasoning of \textsc{LM-Planner} with Llama 3.1 405B in generating missing ablations for Accelerating Retrieval-augmented Language Model Serving with Speculation.}
\label{fig:accelerating_reasoning_llama}
}

{
\begin{reviewerpromptbox}{Reasoning - Qwen 3.5}
After analyzing the paper, I notice that while the authors conducted extensive experiments to validate their method, there are some important ablations missing that could provide deeper insights into the effectiveness of RaLMSpec. Here are two key missing ablations:

1. Cache Size Impact: While the paper uses a local cache for speculative retrieval, there is no systematic study of how the cache size affects performance. The authors only test prefetch sizes of 20 and 256 documents, but don't explore the fundamental trade-off between cache size, hit rate, and memory overhead. This is crucial since the cache is a core component of their speculation mechanism.

2. Verification Batch Size: The paper relies heavily on batched verification but doesn't analyze how different batch sizes affect the overall performance across different retrievers. While they show that batched retrieval is more efficient than sequential retrieval, they don't explore the optimal batch size for different scenarios. This is important since the batch size directly impacts both latency and resource utilization.

These ablations would help better understand the method's sensitivity to key hyperparameters and provide practical guidance for deployment.

The metrics used in the paper for evaluating performance are primarily speed-up ratio and latency (in seconds), so I'll include these in the suggested ablations.
\end{reviewerpromptbox}
\captionsetup{type=figure}
\captionof{figure}{The reasoning of \textsc{LM-Planner} with Qwen 3.5 in generating missing ablations for Accelerating Retrieval-augmented Language Model Serving with Speculation.}
\label{fig:accelerating_reasoning_Qwen}
}

\paragraph{Unpacking SDXL Turbo: Interpreting Text-to-Image Models with Sparse Autoencoders~\cite{surkov2024unpacking}.} The paper studies SDXL Turbo's internal representations using sparse autoencoders (SAEs) to learn interpretable features. Both models struggle at identifying the ablations already presented in the paper, as the paper does not present them in a dedicated form. Llama's reasoning (\Cref{fig:unpacking_reasoning_llama}) describes the general method and then proposes two ablations: one aligns with reviewer suggestions and the other explores hyperparameters of the SAE, which could be interpreted as an ablation. Qwen (\Cref{fig:unpacking_reasoning_Qwen}), in contrast, identifies various analyses in the paper but mistakenly treats them as ablations and then suggests additional hyperparameter-focused experiments rather than altering components to produce insights about the method. 

{
\begin{reviewerpromptbox}{Reasoning - Llama 3.1 405B}
The paper "Unpacking SDXL Turbo: Interpreting Text-to-Image Models with Sparse Autoencoders" presents a study on the interpretability of text-to-image models, specifically the SDXL Turbo model. The authors use sparse autoencoders (SAEs) to learn interpretable features from the intermediate representations of the SDXL Turbo model. They find that the learned features are highly interpretable, causally influence the generation process, and reveal specialization among the blocks of the model.

The authors also present a case study on the most active features for a given prompt and find that the features learned by the SAEs correspond to specific objects, textures, and styles in the generated images. They also find that the features learned by the SAEs can be used to manipulate the generated images in a controlled manner.

The paper also discusses the limitations of the study, including the fact that the SAEs are trained on a limited dataset and that the features learned by the SAEs may not be universally applicable.

Overall, the paper presents a promising approach to interpreting text-to-image models and demonstrates the potential of SAEs in revealing the internal workings of these models.

Based on the paper, I suggest two missing ablation studies:

1. Ablating the number of layers in the SAEs: The authors use a fixed number of layers in the SAEs, but it would be interesting to see how the performance of the SAEs changes when the number of layers is increased or decreased.

2. Ablating the type of activation function used in the SAEs: The authors use the ReLU activation function in the SAEs, but it would be interesting to see how the performance of the SAEs changes when different activation functions are used.

These ablation studies could provide further insights into the workings of the SAEs and the SDXL Turbo model, and could potentially lead to improvements in the performance of the SAEs.
\end{reviewerpromptbox}
\captionsetup{type=figure}
\captionof{figure}{The reasoning of \textsc{LM-Planner} with Llama 3.1 405B in generating missing ablations for Unpacking SDXL Turbo: Interpreting Text-to-Image Models with Sparse Autoencoders.}
\label{fig:unpacking_reasoning_llama}
}

{
\begin{reviewerpromptbox}{Reasoning - Qwen 3.5}
The paper investigates using Sparse Autoencoders (SAEs) to interpret SDXL Turbo's internal representations. The authors train SAEs on updates performed by transformer blocks within the model's denoising U-net and analyze the learned features.

The paper already includes several analyses:

1. Qualitative analysis of learned features through visualization

2. Quantitative analysis of feature interpretability using CLIP similarity

3. Causality analysis comparing empty prompt interventions with feature captions

4. Sensitivity analysis checking if features activate on relevant contexts

5. Analysis of texture and color properties of different transformer blocks

However, I notice two important missing ablation studies:

1. The choice of which intermediate representations to analyze - the authors chose to train SAEs on the updates performed by transformer blocks, but did not justify this choice compared to alternatives like training on the full residual stream or individual attention heads.

2. The architecture of the SAE itself - while they followed standard practices, they did not ablate key architectural choices like the number of features (nf) or sparsity level (k) to demonstrate their impact on feature interpretability and specificity.

These ablations would help validate the key methodological choices and strengthen the paper's conclusions.
\end{reviewerpromptbox}
\captionsetup{type=figure}
\captionof{figure}{The reasoning of \textsc{LM-Planner} with Qwen 3.5 in generating missing ablations for Unpacking SDXL Turbo: Interpreting Text-to-Image Models with Sparse Autoencoders.}
\label{fig:unpacking_reasoning_Qwen}
}

\paragraph{Growth Inhibitors for Suppressing Inappropriate Image Concepts in Diffusion Models~\cite{chen2025growth}.} The paper introduces GIE, a method for eliminating NSFW content in diffusion models by injecting reweighted "growth inhibitors" into attention maps, guided by a suppression-scale adapter. Llama's reasoning (\Cref{fig:growth_reasoning_llama}) does not analyze the ablations already performed in the paper, which makes it easy for the model to generate ablation ideas from scratch; in some cases, these suggestions hit missing ablations that the reviewers also suggested. Qwen (\Cref{fig:growth_reasoning_Qwen}), in contrast, correctly identifies the experiments already performed by the authors, although these are not explicitly labeled as ablations, and additionally proposes two missing ablations from the experimental section that are genuinely relevant, one of them also appears in the reviewer suggestion.

{
\begin{reviewerpromptbox}{Reasoning - Llama 3.1 405B}
The paper "Growth Inhibitors for Suppressing Inappropriate Image Concepts in Diffusion Models" presents a novel approach to suppressing inappropriate features in the image space using growth inhibitors without fine-tuning. The method, called GIE, injects growth inhibitors into the attention map group of the prompt to guide inappropriate features into appropriate ones during the diffusion process. The paper also proposes a scheme for training an adapter to infer the suppression scale of GIE based on the intermediate values of the cross-attention layers.

The paper evaluates the performance of GIE against several baselines for concept erasure in diffusion models and demonstrates its effectiveness in erasing NSFW content, styles, and specific common objects with little effect on unrelated concepts. The paper also shows that GIE preserves the quality and semantics of the generated images.

To further investigate the effectiveness of GIE, we suggest two missing ablation studies:

1. **Ablation of the adapter**: The paper proposes using an adapter to infer the suppression scale of GIE. However, it would be interesting to see how GIE performs without the adapter, using a fixed suppression scale instead. This ablation study would help to understand the importance of the adapter in the GIE framework.

2. **Ablation of the attention map group injection position**: The paper injects the growth inhibitor into the attention map group at the position preceding m[EOT]. However, it would be interesting to see how GIE performs when injecting the growth inhibitor at different positions in the attention map group. This ablation study would help to understand the importance of the injection position in the GIE framework.

These ablation studies would provide further insights into the GIE framework and help to identify potential areas for improvement.

\end{reviewerpromptbox}
\captionsetup{type=figure}
\captionof{figure}{The reasoning of \textsc{LM-Planner} with Llama 3.1 405B in generating missing ablations for Growth Inhibitors for Suppressing Inappropriate Image Concepts in Diffusion Models.}
\label{fig:growth_reasoning_llama}
}

{
\begin{reviewerpromptbox}{Reasoning - Qwen 3.5}
After analyzing the paper, I notice that the authors have conducted several ablation studies, including:
1. Testing different positions for injecting growth inhibitors (Appendix B)

2. Comparing fixed suppression scales vs adaptive scales via the adapter (Appendix C)

3. Testing impact of prompt lengths (Appendix E)

4. Testing effectiveness across different Stable Diffusion models (Appendix G)

However, there are two critical missing ablation studies that would help better understand the method's key components:

1. The most important missing ablation is on the feature extraction process. The authors extract features from the attention maps corresponding to target tokens, but don't validate if this is the optimal approach. Alternative feature extraction methods could potentially improve performance.

2. The second important missing ablation is on the adapter architecture. The authors use a 2-layer MLP with specific dimensions (64->32), but don't justify this choice. Testing different architectures could reveal if this is optimal.

These ablations would help understand the contribution of key design choices to the method's performance.
\end{reviewerpromptbox}
\captionsetup{type=figure}
\captionof{figure}{The reasoning of \textsc{LM-Planner} with Qwen 3.5 in generating missing ablations for Growth Inhibitors for Suppressing Inappropriate Image Concepts in Diffusion Models.}
\label{fig:growth_reasoning_Qwen}
}

\paragraph{UniPAD: A Universal Pre-training Paradigm for Autonomous Driving~\cite{yang2024unipad}.} UniPAD introduces a self-supervised learning paradigm applying 3D volumetric differentiable rendering in autonomous driving, to facilitating the reconstruction of continuous 3D shape structures and the intricate appearance characteristics of their 2D projections. Llama's reasoning (\Cref{fig:unipad_reasoning_llama}) ignores the ablations already present in the paper, continuing the trend from previous papers where it starts from scratch. This makes it easy for the model to suggest potentially meaningful ablations. Its reasoning is not fully coherent: for example, it proposes replacing components in the reasoning trace but then suggests removing them in the actual ablation plan. Qwen (\Cref{fig:unipad_reasoning_Qwen}) correctly identifies the ablations performed in the paper but then focuses on low-level hyperparameter variations, such as voxel resolution and loss weighting, rather than ablations that test major methodological components; these experiments are feasible but do not provide high-level insights about the method.

{
\begin{reviewerpromptbox}{Reasoning - Llama 3.1 405B}
The paper "UniPAD: A Universal Pre-training Paradigm for Autonomous Driving" presents a novel self-supervised learning paradigm for 3D point cloud data. The method uses 3D differentiable rendering to reconstruct the missing geometry on the projected 2D depth image. The paper conducts extensive experiments on various downstream 3D tasks and achieves state-of-the-art results.

Upon reviewing the paper, I noticed that the authors did not conduct ablation studies on the effect of the 3D differentiable rendering module and the impact of the masking strategy on the performance of the model. Therefore, I suggest two missing ablation studies:

1. Ablation study on the 3D differentiable rendering module: This study would involve removing or replacing the 3D differentiable rendering module with a different rendering method to assess its impact on the model's performance.

2. Ablation study on the masking strategy: This study would involve varying the masking ratio or using different masking strategies to assess their impact on the model's performance.

These ablation studies would provide valuable insights into the effectiveness of the 3D differentiable rendering module and the masking strategy, which are crucial components of the UniPAD paradigm.
\end{reviewerpromptbox}
\captionsetup{type=figure}
\captionof{figure}{The reasoning of \textsc{LM-Planner} with Llama 3.1 405B in generating missing ablations for UniPAD: A Universal Pre-training Paradigm for Autonomous Driving.}
\label{fig:unipad_reasoning_llama}
}

{
\begin{reviewerpromptbox}{Reasoning - Qwen 3.5}
After analyzing the paper, I notice that the authors have conducted several ablation studies, including:
1. Masking ratio experiments

2. Decoder depth and width studies

3. Different rendering techniques comparison

4. Ray sampling strategies evaluation

5. Feature projection analysis

6. Pre-trained components analysis

However, there are two critical components that lack proper ablation studies:

1. The volumetric representation resolution (X×Y×Z): The paper uses a fixed voxel resolution of 180×180×5, but doesn't investigate how this choice affects performance. This is crucial since the voxel resolution directly impacts both the computational cost and the representation quality.

2. The rendering loss components: The paper uses a combination of RGB and depth losses (Eq. 4), but doesn't investigate the relative importance of each loss term or explore different loss weightings ($\lambda_\text{RGB}$ and $\lambda_\text{depth}$). Understanding the contribution of each loss component is essential for the method's effectiveness.

These missing ablations would help better understand the method's sensitivity to these important design choices and potentially improve its performance.
\end{reviewerpromptbox}
\captionsetup{type=figure}
\captionof{figure}{The reasoning of \textsc{LM-Planner} with Qwen 3.5 in generating missing ablations for UniPAD: A Universal Pre-training Paradigm for Autonomous Driving.}
\label{fig:unipad_reasoning_Qwen}
}

\subsection{Human Evaluation} \label{app:human_evaluation_settings}

This section describes the instructions and processing of the human evaluation results in~\Cref{sec:human_eval}.

To conduct the evaluation, we begin by randomly selecting 15 test-set papers, 5 from each broad category: Computer Vision, NLP, and General Machine Learning, and truncating each to include all sections up to and including the method section. We recruited 15 participants currently enrolled in a PhD in machine learning. Each participant is given a questionnaire, where he/she chooses one paper based on their expertise and generated an ablation plan following \researcherbench{}'s formulation (\Cref{fig:human_eval}) using the truncated paper. Participants were instructed to spend at least one hour on the task and explicitly not to use any generative AI tools to ensure a fair comparison.

We manually inspect all submissions and exclude 5 responses where participants did not demonstrate a clear understanding of the task or the paper, for example by proposing general experiments instead of ablations or by leaving placeholder content in required fields. The remaining 10 submissions are evaluated against the GT ablations using the same matching and scoring criteria applied to model generations. We report the same evaluation metrics as in \researcherbench{} and compare human performance to the best-performing model on this task, Qwen 3.5, evaluated on the same subset of papers.

The responses of all participants are supplied as part of the supplementary material. The papers that were part of the survey by category are given below.

\textbf{Computer Vision:}
\begin{itemize}
    \item Real-Time High-Resolution Background Matting~\cite{lin2020realtimehighresolutionbackgroundmatting}.
    \item SpiderMatch: 3D Shape Matching with Global Optimality and Geometric Consistency~\cite{roetzer2024spidermatch}.
    \item Learning High Fidelity Depths of Dressed Humans by Watching Social Media Dance Videos~\cite{Jafarian_2021_CVPR_TikTok}.
    \item Good Visual Guidance Makes A Better Extractor: Hierarchical Visual Prefix for Multimodal Entity and Relation Extraction~\cite{chen-etal-2022-good}.
    \item Image Processing GNN: Breaking Rigidity in Super-Resolution~\cite{Tian_2024_CVPR}.
\end{itemize}

\textbf{NLP:}
\begin{itemize}
    \item DisCoDisCo at the DISRPT2021 Shared Task: A System for Discourse Segmentation, Classification, and Connective Detection~\cite{gessler-etal-2021-discodisco}.
    \item Preserving Commonsense Knowledge from Pre-trained Language Models via Causal Inference~\cite{zheng2023preservingcommonsenseknowledgepretrained}.
    \item Assisting in Writing Wikipedia-like Articles From Scratch with Large Language Models~\cite{shao2024assistingwritingwikipedialikearticles}.
    \item Differentiable Prompt Makes Pre-trained Language Models Better Few-shot Learners~\cite{zhang2022differentiable}.
    \item Learning Performance-Improving Code Edits~\cite{shypula2024learningperformanceimprovingcodeedits}.
\end{itemize}

\textbf{General Machine Learning:}
\begin{itemize}
    \item Self-Composing Policies for Scalable Continual Reinforcement Learning~\cite{malagón2025selfcomposingpoliciesscalablecontinual}.
    \item APT: Adaptive Pruning and Tuning Pretrained Language Models for Efficient Training and Inference~\cite{zhao2024apt}.
    \item LCA-on-the-Line: Benchmarking Out-of-Distribution Generalization with Class Taxonomies~\cite{shi2024lcaonthelinebenchmarkingoutofdistributiongeneralization}.
    \item A Generalization of Transformer Networks to Graphs~\cite{dwivedi2021generalizationtransformernetworksgraphs}.
\end{itemize}

{
\begin{researcherpromptbox}{Human Evaluation - Questionnaire}
We are building a new benchmark to evaluate the scientific research capabilities of language models (LLMs). Our current focus is on asking LLM agents to plan the ablation study of a paper, given its method section.

We ask you to create ablation study plan given a paper's method section only.

Please follow the exact instructions to make our evaluation valid and reliable.

It should take about approximately 1 hour to participate - mostly by reading and understanding the paper.

Thanks for your help!

\smallskip
\textbf{Question 1:}

Choose your expertise domain:

\begin{enumerate}
    \item Computer Vision.
    \item NLP.
    \item General Machine Learning.
\end{enumerate}

\smallskip
\textbf{Question 2:}

Please choose one paper with the following rules:
\begin{itemize}
    \item A paper the you are NOT familiar with.
    \item Do not search the paper online, use the provided link to get the paper PDF (which includes all the section up to and including the method section).
\end{itemize}

List of papers (please open the link now of the paper you choose):

[List of papers is presented here, per category]

\smallskip
\textbf{Questions 3-7:}

Your task now is to generate UP TO 5 ablations (ordered by relevance - from high to low) of the paper you have chosen. 

Keep in mind that ablation studies aim to attribute the method's performance to its major components.

\textit{Please do NOT use any GenAI in your ablation studies plan.}

Each ablation in the plan should contain:
\begin{enumerate}
    \item \textbf{Ablated part/component:} A high-level description of the part or component of the method that you think should be ablated.

    \item \textbf{Action:} The action to perform on the ablated part (one of: Remove, Replace, Add). If the action is Replace or Add, please include in the 'replacement' a list of all replacements/additions you think are relevant to test.

    \item \textbf{Replacement:} Include this field only if the action is Replace or Add. Provide a list of all the replacements you think are relevant to test.
    \item \textbf{Metrics:} A list of metrics to evaluate the ablation study.
    \item \textbf{Rationale:} Why you think this ablation is relevant in the context of the paper?
\end{enumerate}

After each ablation you add, you have the option to terminate the survey or to suggest another ablation.

\end{researcherpromptbox}
\captionsetup{type=figure}
\captionof{figure}{The instructions for the human evaluation survey.} \label{fig:human_eval}
}

\newpage

\section{Prompts} \label{app:prompts}
This section provides the prompt for all the LM systems in \benchmark{} including: the two planners and two judges of \researcherbench{} and the two planners and two judges of \reviewerbench{}.

\subsection{\researcherbench{}}

\subsubsection{\textsc{LMJudge}}

{
\begin{researcherpromptbox}{System Prompt}
\textbf{SETTING:} You are an autonomous computer science researcher, an expert at analyzing machine learning papers and their ablation studies.

\smallskip
You are given a matching task between two sets of ablations: those presented in a research paper and those suggested in an ablation plan, but you do not know which set corresponds to the paper and which corresponds to the plan.  
A match means: The experiment described by an ablation in side~A is allowed or included as a possible option within an ablation in side~B (and vice versa).

\smallskip
You need to format your output using two fields: discussion and predictions.  
The discussion field should contain a clear explanation of your reasoning, and the predictions field should contain the final output.  
The predictions field should be in a strict JSONL format, with each line containing a JSON object representing the final output.

\smallskip
Each JSON object should have the following fields:

\smallskip
1. ``name\_in\_A'': name(s) of the matching ablations in the ablations given in side~A under \texttt{<ablations\_in\_A></ablations\_in\_A>}.

\smallskip
2. ``name\_in\_B'': name(s) of the matching ablations in the ablations given in side~B under \texttt{<ablations\_in\_B></ablations\_in\_B>}.

\smallskip
Your output should always include \_one\_ discussion and \_one\_ predictions field EXACTLY as in the following example:

\smallskip
<discussion>

The ablations in side~A contain three different ablations, and the ablations in side~B contain four different ablations. We analyze each of them separately.

\smallskip
Side~A:

\smallskip
1. Ablation~A: This ablation is about the model architecture. The ablations in side~B contain a similar ablation (Ablation~X) with the same ablated part and action, so we consider it a match.

\smallskip
2. Ablation~B: This ablation is about the training data. The ablations in side~B contain a similar ablation with a different name (Ablation~Y), but the ablated part, action, and replacement content match exactly, so we consider it a match.

\smallskip
3. Ablation~C: This ablation is about the evaluation metric. The ablations in side~B do not contain any similar ablation, so we cannot consider it a match.

\smallskip
Side~B:

\smallskip
1. Ablation~X: was matched to Ablation~A from side~A.

\smallskip
2. Ablation~Y: was matched to Ablation~B from side~A.

\smallskip
3. Ablation~Z: there is no matching ablation in side~A.

\smallskip
4. Ablation~W: there is no matching ablation in side~A.

</discussion>

\smallskip
<predictions>

\{"name\_in\_A": "Ablation A", "name\_in\_B": "Ablation X"\}

\{"name\_in\_A": "Ablation B", "name\_in\_B": "Ablation Y"\}

\{"name\_in\_A": "Ablation C", "name\_in\_B": null\}

\{"name\_in\_A": null, "name\_in\_B": "Ablation Z"\}

\{"name\_in\_A": null, "name\_in\_B": "Ablation W"\}

</predictions>
\smallskip

\end{researcherpromptbox}%
\captionsetup{type=figure}
\captionof{figure}{The system prompt for \textsc{LMJudge} for \researcherbench{}.}
}

{
\begin{researcherpromptbox}{Instance Prompt}

\textbf{SETTING:} We're currently reviewing an ablation studies plan for the research paper \{\{paper\_title\}\}. Here's the research abstract:

\smallskip
\textbf{ABSTRACT:}  
\{\{abstract\}\}

\smallskip
\textbf{INSTRUCTIONS:}  
Below, you will find the ablations performed in the paper and the suggested ablation plan.  
We do not provide any information which ablations belong to which source, so you need to carefully analyze each ablation in both paper and plan and determine if there is a match.  
Your task is to find a match between ablations in side~A and ablations in side~B, and vice versa.

\smallskip
A match is considered valid only if all of the following are true:

\smallskip
1. The ``ablated\_part'' (i.e., the component or mechanism being ablated) must refer to the same component.

2. The ``action'' in one ablation must be explicitly allowed by an ablation in the other side.

3. If the action is REPLACE or ADD, the ``replacement'' content must have at least one valid option match.

\smallskip
Examples:

\smallskip
- REMOVE X → REPLACE X with [remove, Y]                MATCH  

- REPLACE X with Y → REMOVE X                          NO MATCH  

- REMOVE X → REMOVE X                                  MATCH  

- REPLACE X with [Y, Z] → REPLACE X with [Z, W]       MATCH  

- ADD Y to X → ADD [Y, Z] to X                        MATCH  

- REPLACE (X+Y) with [X, Y, Z] → [REMOVE X, REMOVE Y] MATCH

\smallskip
Be conservative: if it is unclear whether one ablation allows the other ablation, do NOT match.

\smallskip
The match can contain multiple ablations from either side, as long as the match meets the criteria above.  
For example, if an ablation in side~A is split into two ablations in side~B, but both of them together meet the matching criteria, then it is still considered a valid match.

\smallskip
\textbf{What You Need to Do:}  

\smallskip
Fill in the following fields for each entry:

\smallskip
\{"name\_in\_A": [ name of the ablation in side~A OR a list of ablation names in side~A OR null if there is no matching ablation for the ablation in side~B ], "name\_in\_B": [ name of the matching ablation from side~B OR a list of ablation names in side~B OR null if there is no matching ablation for the ablation in side~A ]\}

\smallskip
\textbf{GENERAL IMPORTANT TIPS:}  

\smallskip
1. The ``name\_in\_A'' and ``name\_in\_B'' should match exactly to the ablation names in side~A and side~B, respectively.

2. You must go over ALL the ablations in side~A and all ablations in side~B, \_each one of them should be in your prediction\_.

3. It is OK if one ablation from side~A is split into multiple ablations in side~B, or merged several side~A ablations into one — as long as the matching criteria (per ablation in side~A) are met, the match is still valid. This holds to the other direction as well (side~B to side~A).

\smallskip
\textbf{STRATEGY:}  

\smallskip
1. For each ablation in side~A, you should decide based on the criteria above if there is a valid match in side~B, and vice versa.

2. Please make sure to include in the discussion field the reasoning behind your decision for each ablation.

3. After that, you should have the information to fill in the ``name\_in\_B'' / ``name\_in\_A'' fields for each ablation in side~A and side~B, respectively. If the ablation is considered a match to one (or more) of the ablations in the other side, you should fill the appropriate ablation(s) in ``name\_in\_B'' / ``name\_in\_A'', otherwise it should remain null.

\smallskip
Here are the ablations in side~A and side~B:

\smallskip
<ablations\_in\_A>  
\{\{side\_A\}\}  
</ablations\_in\_A>

\smallskip
<ablations\_in\_B>  
\{\{side\_B\}\}  
</ablations\_in\_B>

\smallskip
**Remember to include all the ablations in side~A and side~B in your predictions!**

\smallskip
BELOW IS AN EXAMPLE THAT WILL HELP YOU UNDERSTAND THE FORMATTING AND THE TASK BETTER. PLEASE USE IT FOR YOUR REFERENCE ONLY AND DO NOT REPEAT IT IN YOUR FINAL ANSWER.

\smallskip
Example:

\smallskip
<ablations\_in\_A>  
\{"name": "HybrIK Variant Comparison", "ablated\_part": "Adaptive HybrIK mechanism", "action": "REPLACE", "replacement": ["Naive HybrIK (using original parent joints)", "Adaptive HybrIK (using reconstructed parent joints)", "Iterative global optimization (non-differentiable baseline)"], "metrics": ["MPJPE (mm)", "PVE (mm)", "Error accumulation along kinematic tree"]\}  

\{"name": "Shape Parameter Study", "ablated\_part": "SMPL shape parameter prediction", "action": "REPLACE", "replacement": ["Mean shape parameters", "PCA-based shape space reduction", "Direct vertex offset prediction"], "metrics": ["MPJPE (mm)", "PVE (mm)", "Shape error metrics"]\}  

\{"name": "Twist-Swing Ablation", "ablated\_part": "Twist-and-swing decomposition of rotations", "action": "REPLACE", "replacement": ["Direct rotation regression (3-DoF) as in previous works", "Only swing component (2-DoF)", "Only twist component (1-DoF)"], "metrics": ["MPJPE (mm)", "PVE (mm)", "Reconstruction error"]\}  

\{"name": "3D Keypoint Estimation Study", "ablated\_part": "3D keypoint estimation component", "action": "REPLACE", "replacement": ["Direct regression without volumetric representation", "Different heatmap resolutions", "Alternative backbone architectures"], "metrics": ["MPJPE (mm)", "PVE (mm)", "Per-joint accuracy"]\}  

\{"name": "Twist Angle Prediction", "ablated\_part": "Twist angle prediction network", "action": "REPLACE", "replacement": ["Fixed twist angles (0 degrees)", "Constrained range predictions", "Multi-head prediction network"], "metrics": ["MPJPE (mm)", "PVE (mm)", "Physical plausibility score"]\}  

</ablations\_in\_A>

\smallskip
<ablations\_in\_B>  
\{"name": "Analysis of the twist rotation", "ablated\_part": "Twist angles in the twist-and-swing decomposition", "action": "REPLACE", "replacement": ["Random values in [-$\pi$, $\pi$]", "Zero twist angles", "Network-estimated twist angles"], "metrics": ["Mean error of reconstructed 24 SMPL joints", "Mean error of 14 LSP joints", "Mean error of body mesh", "Mean error of twist angle"]\}  

\{"name": "Error correction capability of HybrIK", "ablated\_part": "Apply different algorithms on the predicted pose and compare it to HybrIK", "action": "REPLACE", "replacement": ["SMPLify"], "metrics": ["MPJPE"]\}  

\{"name": "Effect of shape parameters ($\beta$)", "ablated\_part": "Shape parameters ($\beta$) in SMPL model", "action": "REPLACE", "replacement": ["Ground-truth $\beta$", "Zero $\beta$"], "metrics": ["MPJPE", "PVE"]\}  

\{"name": "Robustness of HybrIK to noise", "ablated\_part": "Comparison between Naive HybrIK to Adaptive HybrIK with different input joints", "action": "REPLACE", "replacement": ["GT joints", "10mm noise", "20mm noise", "30mm noise"], "metrics": ["MPJPE"]\}  

</ablations\_in\_B>

\smallskip
<discussion>  

Let's go over all ablations in side~A and side~B to determine the matches based on the criteria provided.

\smallskip
Side~A:

\smallskip
1. HybrIK Variant Comparison: This ablation replaces Adaptive HybrIK mechanism with several variants. In side~B, the ablation "Robustness of HybrIK to noise" deals with the adaptive HybrIK, it does replace it with one variant (GT joints) that matches side~A variant (Naive HybrIK (using original parent joints)). Therefore, there is a match.

\smallskip
2. Shape Parameter Study: This ablation replaces SMPL shape parameter prediction with different methods. In side~B, "Effect of shape parameters ($\beta$)" replaces shape parameters with ground-truth and zero $\beta$, which does not align with the replacement criteria. Therefore, no match.

\smallskip
3. Twist-Swing Ablation: This ablation replaces the twist-and-swing decomposition with other methods. In side~B, "Analysis of the twist rotation" is the closest one, though it focuses on twist angles specifically. Therefore, no match.

\smallskip
4. 3D Keypoint Estimation Study: This ablation replaces the 3D keypoint estimation component with various alternatives. None of the ablations in side~B address this component, so no match.

\smallskip
5. Twist angle prediction: This ablation replaces the twist angle prediction network with different methods. In side~B, "Analysis of the twist rotation" again focuses on twist angles with one replacement (zero twist angles) that matches. Therefore, we have a match here.

\smallskip
Side~B:

\smallskip
1. Analysis of the twist rotation: was matched to "Twist angle prediction" from side~A.

\smallskip
2. Error correction capability of HybrIK: There is no matching ablation in side~A, as none of the ablations in side~A address error correction on the predicted pose.

\smallskip
3. Effect of shape parameters ($\beta$): There is no matching ablation in side~A, the closest one is "Shape Parameter Study", but the replacements do not align.

\smallskip
4. Robustness of HybrIK to noise: was matched to "HybrIK Variant Comparison" from side~A.  

</discussion>

\smallskip
<predictions>  

\{"name\_in\_A": "HybrIK Variant Comparison", "name\_in\_B": "Robustness of HybrIK to noise"\}  

\{"name\_in\_A": "Shape Parameter Study", "name\_in\_B": null\}  

\{"name\_in\_A": "Twist-Swing Ablation", "name\_in\_B": null\}  

\{"name\_in\_A": "3D Keypoint Estimation Study", "name\_in\_B": null\} 

\{"name\_in\_A": "Twist Angle Prediction", "name\_in\_B": "Analysis of the twist rotation"\}  

\{"name\_in\_A": null, "name\_in\_B": "Error correction capability of HybrIK"\}  

\{"name\_in\_A": null, "name\_in\_B": "Effect of shape parameters ($\beta$)"\}  

</predictions>
\smallskip

\end{researcherpromptbox}
\captionsetup{type=figure}
\captionof{figure}{The task instance prompt for \textsc{LMJudge} for \researcherbench{}.}
}

\subsubsection{\textsc{AgentJudge}}

{
\begin{researcherpromptbox}{System Prompt}
\textbf{SETTING:} You are an autonomous computer science researcher, and you're working directly in the command line of a Linux container with a special interface.

\smallskip
You are given a matching task between two sets of ablations: those presented in a research paper and those suggested in an ablation plan, but you do not know which set corresponds to the paper and which corresponds to the plan.  
A match means: The experiment described by ablation in side~A is allowed or included as a possible option within ablation in side~B (and vice versa).

\smallskip
The special interface consists of a file editor that shows you \{\{WINDOW\}\} lines of a file at a time.  
In addition to typical bash commands, you can also use specific commands to help you navigate within files.  
To call a command, you need to invoke it with a function call/tool call.

\smallskip
\textbf{RESPONSE FORMAT:}  
Your shell prompt is formatted as follows:  

\texttt{(Open file: <path>)}

\texttt{(Current directory: <cwd>)  }

\texttt{bash-\$}

\smallskip
First, you should \_always\_ include a general thought about what you're going to do next.  
Then, for every response, you must include exactly \_ONE\_ tool call/function call.

\smallskip
Remember, you should always include a \_SINGLE\_ tool call/function call and then wait for a response from the shell before continuing with more discussion and commands. Everything you include in the DISCUSSION section will be saved for future reference.  
If you'd like to issue two commands at once, PLEASE DO NOT DO THAT! Please instead first submit just the first tool call, and then after receiving a response you'll be able to issue the second.  
Note that the environment does NOT support interactive session commands (e.g. python, vim), so please do not invoke them.
\smallskip

\end{researcherpromptbox}
\captionsetup{type=figure}
\captionof{figure}{The system prompt for SWE-agent judge for \researcherbench{}.} \label{fig:agent_system_prompt}
}

{
\begin{researcherpromptbox}{Instance Prompt}
\textbf{SETTING:} You are in an empty repository root directory `/repo`.  
We're currently want to review an ablation studies plan for the research paper \{\{paper\_title\}\}. Here's the research abstract:  

\smallskip
\textbf{ABSTRACT:}  
\{\{abstract\}\}

\smallskip
\textbf{INSTRUCTIONS:}  
Below, you will find the ablations performed in the paper and the suggested ablation plan.  
We do not provide any information which ablations belong to which source, so you need to carefully analyze each ablation in both paper and plan and determine if there is a match.  
Your task is to find a match between ablations in side~A and ablations in side~B, and vice versa.

\smallskip
A match is considered valid only if all of the following are true:

1. The `ablated\_part` (i.e., the component or mechanism being ablated) must refer to the same component.

2. The `action` in one ablation must be explicitly allowed by an ablation in the other side.

3. If the action is REPLACE or ADD, the `replacement` content must have at least one valid option match.

\smallskip
Examples:

- REMOVE X → REPLACE X with [remove, Y]                MATCH  

- REPLACE X with Y → REMOVE X                          NO MATCH  

- REMOVE X → REMOVE X                                  MATCH  

- REPLACE X with [Y, Z] → REPLACE X with [Z, W]       MATCH  

- ADD Y to X → ADD  [Y, Z] to X                        MATCH  

- REPLACE (X+Y) with [X, Y, Z] → [REMOVE X, REMOVE Y]  MATCH

\smallskip
Be conservative: if it is unclear whether one ablation allows the other ablation, do NOT match.  

The match can contain multiple ablations from either side, as long as the match meets the criteria above.  
For example, if an ablation in side~A is split into two ablations in side~B, but both of them together meet the matching criteria, then it is still considered a valid match.

\smallskip
\textbf{What You Need to Do:}  

Create the file `/repo/final\_score.jsonl` using the command `create\_final\_score` and then fill in the following fields for each entry:  

\smallskip

\{"name\_in\_A": [ name of the ablation in side A OR a list of ablation names in side A OR null if there is no matching ablation for the ablation in side B ], 
 "name\_in\_B": [ name of the matching ablation from side B OR a list of ablation names in side B OR null if there is no matching ablation for the ablation in side A ]\}

\smallskip
\textbf{GENERAL IMPORTANT TIPS:}  

1. The file should be initialized using the command `create\_final\_score`, after that you just need to edit it and fill in the `name\_in\_B` or `name\_in\_A` (where applicable) field.  

2. The `name\_in\_A` and `name\_in\_B` should match exactly to the ablation names in side~A and side~B, respectively.  

3. You must go over ALL the ablations in side~A and all ablations in side~B, \_each one of them should be in your prediction\_.  

4. It is OK if one ablation from side~A is split into multiple ablations in side~B, or merged several side~A ablations into one — as long as the matching criteria (per ablation in side~A) are met, the match is still valid. This holds to the other direction as well (side~B to side~A).  

5. Please note that all information is in the following text and not in any file or directories in the environment, so you should base your decision only using the information below.  

6. Any operations on the environment besides creating the `/repo/final\_score.jsonl` file and editing it are not necessary.

\smallskip
\textbf{STRATEGY:}  

1. First call the `create\_final\_score` command to create the file.  

2. Then, for each ablation in side~A, you should decide based on the criteria above if there is a valid match in side~B, and vice versa.  

3. After that, you should have the information to fill in the `name\_in\_B` / `name\_in\_A` fields for each ablation in side~A and side~B, respectively. If the ablation is considered a match to one (or more) of the ablations in the other side, you should fill the appropriate ablation(s) in `name\_in\_B` / `name\_in\_A`, otherwise it should remain null.  

4. Submit your final score using the `submit` command.  

PLEASE NOTE THAT THE SUBMISSION FILE NEEDS TO BE INITIALIZED USING THE `create\_final\_score` COMMAND.  
When you are done, run the `submit` command to confirm.  
You have access to the terminal session, so take your time and be precise.

\smallskip
Here are the ablations in side~A and side~B:

<ablations\_in\_A>  
\{\{side\_A\}\}  
</ablations\_in\_A>

<ablations\_in\_B>  
\{\{side\_B\}\}  
</ablations\_in\_B>

\smallskip
\textbf{Remember to include all the ablations in side~A and side~B in your predictions!}

\smallskip
\texttt{(Open file: \{\{open\_file\}\})  }

\texttt{(Current directory: \{\{working\_dir\}\})  }

\texttt{bash-\$}

\end{researcherpromptbox}
\captionsetup{type=figure}
\captionof{figure}{The task instance prompt for SWE-agent judge for \researcherbench{}.}
}

\subsubsection{\textsc{LM-Planner}}

{
\begin{researcherpromptbox}{System Prompt}
\textbf{SETTING:} You are an autonomous computer science researcher, an expert at analyzing machine learning papers and their ablation studies.

You need to format your output using two fields; discussion and predictions.
The discussion field should contain a clear explanation of your reasoning, and the predictions field should contain the final output.
The predictions field should be in a strict JSONL format, with each line containing a JSON object representing the final output.

\smallskip
Each JSON object should have the following fields:

1. "name": name of the ablation experiment.

2. "ablated\_part": high-level description of the part of the method in the research that you want to ablate.

3. "action": the action you want to take on the ablated part (REMOVE, REPLACE, ADD). If the action is REPLACE or ADD please add a field named "replacement" and specify a list of possible replacements/additions, for example if you want to change the value of a parameter please specify a list of values to test.

4. "metrics": a list of metrics to report of the ablation experiment. Please pay special attention and use the metrics that are also used in the paper.

Your output should always include \_one\_ discussion and \_one\_ predictions field EXACTLY as in the following example:

\smallskip

<discussion>

Output here the step by step reasoning of the ablation plan you suggest.

</discussion>

\smallskip
<predictions>

\{"name": "Ablation A", "ablated\_part": "description of the ablated part", "action": "REMOVE", "metrics": ["metric1", "metric2"]\}

\{"name": "Ablation B", "ablated\_part": "description of the ablated part", "action": "REPLACE", "replacement": ["replacement1", "replacement2"], "metrics": ["metric3"]\}

</predictions>
\end{researcherpromptbox}
\captionsetup{type=figure}
\captionof{figure}{The system prompt for LM planner for \researcherbench{}.} \label{fig:lm_planner_system_prompt}
}

{
\begin{researcherpromptbox}{Instance Prompt}
We're currently want to suggest ablation studies for the research titled \{\{paper\_title\}\}. Here's the research abstract:

\smallskip
ABSTRACT:

\{\{problem\_statement\}\}

The paper source is provided below, after all of the instructions.

\smallskip
INSTRUCTIONS:
Now, you're going to suggest UP TO \{\{num\_ablations\}\} ablation studies on your own, in a JSONL format.
You need to rank the output ablation studies by their importance, and you should only include the most important ones.
Each suggestion should include the following fields in a separate JSON:

1. "name": name of the ablation experiment.

2. "ablated\_part": high-level description of the part of the method in the research that you want to ablate.

3. "action": the action you want to take on the ablated part (REMOVE, REPLACE, ADD). If the action is REPLACE or ADD please add a field named "replacement" and specify a list of possible replacements/additions, for example if you want to change the value of a parameter please specify a list of values to test.

4. "metrics": a list of metrics to report of the ablation experiment. Please pay special attention and use the metrics that are also used in the paper.

When you're satisfied with your ablation studies plan, you can submit your plan.

\smallskip
GENERAL IMPORTANT TIPS:

1. The paper source is provided below, after all of the instructions.

2. Less is more - don't aim to change completely the method, take the important parts and investigate them.

\smallskip
STRATEGY:

1. Read the paper sections, especially the method section, to understand the method.

2. Look at the main components of the method and think about what would happen if you change them.

3. Keep in mind that ablation studies aim to attribute the method's performance to its major components.

5. Remember that ablation studies does not necessarily mean removing parts of the method, but also changing them.

\smallskip
<paper\_source>

\{\{paper\_source\}\}

</paper\_source>
\end{researcherpromptbox}
\captionsetup{type=figure}
\captionof{figure}{The task instance prompt for LM planner for \researcherbench{}.}
}

\subsubsection{\textsc{Agent-Planner}}

The agent planner system prompt is identical to the agent judge system prompt given in~\Cref{fig:agent_system_prompt}.

{
\begin{researcherpromptbox}{Instance Prompt}
You are in an empty repository root directory `/repo`. The paper TeX source is located in the `/paper` read-only directory.
We're currently want to suggest ablation studies for the research titled \{\{paper\_title\}\}. Here's the research abstract:

\smallskip
ABSTRACT:

\{\{problem\_statement\}\}

\smallskip
INSTRUCTIONS:

Now, you're going to suggest UP TO \{\{num\_ablations\}\} ablation studies on your own, in a JSONL file `/repo/ablations\_plan.jsonl`.
You need to rank the output ablation studies by their importance, and you should only include the most important ones.
Each suggestion should include the following fields in a separate JSON:

1. "name": name of the ablation experiment.

2. "ablated\_part": high-level description of the part of the method in the research that you want to ablate.

3. "action": the action you want to take on the ablated part (REMOVE, REPLACE, ADD). If the action is REPLACE or ADD please add a field named "replacement" and specify a list of possible replacements/additions, for example if you want to change the value of a parameter please specify a list of values to test.

4. "metrics": a list of metrics to report of the ablation experiment. Please pay special attention and use the metrics that are also used in the paper.

\smallskip
Your terminal session has started and you're in the repository's root directory.
You can use any bash commands or the special interface to help you.
The environment is already configured. YOU DON'T NEED TO INSTALL ANYTHING IN THE ENVIRONMENT.
Remember, YOU SHOULD ALWAYS INCLUDE EXACTLY ONE TOOL CALL/FUNCTION CALL PER RESPONSE.
When you're satisfied with your ablation studies plan, you can submit your plan to the code base by simply running the submit command.
Note however that you cannot use any interactive session commands (e.g. python, vim) in this environment, but you can write scripts and run them. E.g. you can write a python script and then run it with the python command.

\smallskip

GENERAL IMPORTANT TIPS:

1. To view the paper you should first run `cd /paper`.

2. Less is more - don't aim to change completely the method, take the important parts and investigate them.

3. Don't try to run extensive operations, remember that we just want to create the plan and to execute it for now.

4. If you run a command and it doesn't work, try running a different command. A command that did not work once will not work the second time unless you modify it!

5. If you open a file and need to get to an area around a specific line that is not in the first 100 lines, say line 583, don't just use the scroll\_down command multiple times. Instead, use the goto 583 command. It's much quicker.

6. Always make sure to look at the currently open file and the current working directory (which appears right after the currently open file). The currently open file might be in a different directory than the working directory! Note that some commands, such as 'create', open files, so they might change the current open file.

\smallskip
STRATEGY:

1. Read the paper sections, especially the method section, to understand the method.

2. Look at the main components of the method and think about what would happen if you change them.

3. Keep in mind that ablation studies aim to attribute the method's performance to its major components.

5. Remember that ablation studies does not necessarily mean removing parts of the method, but also changing them.

(Open file: \{\{open\_file\}\})
(Current directory: \{\{working\_dir\}\})
bash-\$
\end{researcherpromptbox}
\captionsetup{type=figure}
\captionof{figure}{The task instance prompt for SWE-agent planner for \researcherbench{}.}
}

\subsection{\reviewerbench{}}

\subsubsection{\textsc{LMJudge}}

{
\begin{reviewerpromptbox}{System Prompt}

\textbf{SETTING:} You are an autonomous computer science researcher, an expert at analyzing machine learning papers and their ablation studies.

\smallskip
You need to format your output using two fields; discussion and predictions.  
The discussion field should contain a clear explanation of your reasoning, and the predictions field should contain the final output.  
The predictions field should be in a strict JSONL format, with each line containing a JSON object representing the final output.

\smallskip
Each JSON object should have the following fields:

1. "name\_in\_plan": name of the ablation in the plan.  

2. "appears\_in\_review": true/false based on the criteria given to you.

\smallskip
Your output should always include \_one\_ discussion and \_one\_ predictions field EXACTLY as in the following example:

\smallskip
<discussion>

The review contains the following new ablations suggestions:

1. Ablation X: The reviewers explicitly suggests that ablation about the attention component is missing.

2. Ablation Y: The reviewers explicitly suggests that other LoRA configuration should be used for training as part of the ablation study.

The ablation in plan contains three different ablations, we will analyze each of them separately:

1. Ablation A: This ablation is about changing the rank as part of the LoRA configuration. This is a match with Ablation Y suggested by the reviewers.

2. Ablation B: This ablation is about the training data. The reviewers did not suggest any ablation related to the training data, so this is not a match.

3. Ablation C: This ablation is about the evaluation metric. The reviewers did not mention any similar ablation, so we cannot consider it a match.

</discussion>

\smallskip
<predictions>  

\{"name\_in\_plan": "Ablation A", "appears\_in\_review": true\}  

\{"name\_in\_plan": "Ablation B", "appears\_in\_review": false\}  

\{"name\_in\_plan": "Ablation C", "appears\_in\_review": false\}  

</predictions>

\smallskip

\end{reviewerpromptbox}
\captionsetup{type=figure}
\captionof{figure}{The system prompt for LM judge for \reviewerbench{}.}
}

{
\begin{reviewerpromptbox}{Instance Prompt}
\textbf{SETTING:} You are currently reviewing an ablation studies plan for the research paper \{\{paper\_title\}\}. Here's the research abstract:

\smallskip
\textbf{ABSTRACT:}  
\{\{abstract\}\}

\smallskip
The paper source is provided below, after all of the instructions.  
You should interpret the review(s) of the paper in the context of the paper source, which may contain important information about the ablation studies performed in the paper, and can help you better understand the ablation suggestions in the review.

\smallskip
\textbf{INSTRUCTIONS:}  
Below, you will find official review(s) of the paper referring to the ablation studies performed in the paper, alongside the suggested ablation plan.  
Your task is to determine, for each ablation in the suggested ablation plan, whether it appears in the review(s).

\smallskip
An ablation in plan is considered to appear in the review if all of the following are true:

1. The \texttt{ablated\_part} (i.e., the component or mechanism being ablated) matches exactly to one of the ablation suggestions in the review text.  

2. The \texttt{action} in one ablation must match exactly (if it exists in the review text ablation suggestion).  

3. If the action is REPLACE or ADD, the \texttt{replacement} content must match (if it exists in the review text ablation suggestion).  
   \smallskip
   * If the field \texttt{"replacement"} contains multiple valid options, your match is valid if at least one correct replacement is identified in the suggested plan.

\smallskip
Be conservative: if it is unclear whether one ablation matches another, do not match.  
However, because the review text may be incomplete, an ablation is still considered a match if the \texttt{ablated\_part} matches exactly (Criterion 1) and the action or replacement information (Criteria 2 and 3) are not mentioned in the review text.

\smallskip
\textbf{What You Need to Do:}  

Fill in the following fields for each suggested ablation in the plan:

\smallskip

\{"name\_in\_plan": [ name of the ablation in the plan ], "appears\_in\_review": [ true/false based on the criteria above ]\}

\smallskip
\textbf{GENERAL IMPORTANT TIPS:}

1. The \texttt{name\_in\_plan} should match exactly to the ablations in the suggested plan.  

2. You must go over all the ablations in the suggested plan.  

3. Please note that all review information is in the following text, so you should base your decision only using the information below.  

4. You should treat the review \_ONLY FOR NEW ABLATION SUGGESTIONS\_, and not for existing ones or other aspects of the paper or the review.  
   For example, the review text may contain suggestions for other experiments which are NOT ablations.  

5. You are given the paper source — use it to better understand the context of the review and the ablation suggestions.  
   For example, the paper source contains already the ablation studies performed in the paper, which can help you understand better the ablation suggestions in the review.

\smallskip
\textbf{STRATEGY:}

1. You should extract from each review the missing ablation that the reviewer suggests to add.  
   A missing ablation refers to an experiment that the reviewer believes should have been conducted, in which a specific component, module, feature, or design choice is removed, replaced, or altered in order to assess its impact on the model's performance.  
   The review must clearly indicate that such an ablation study is missing or should be added.

2. After that, you should have the information to fill in the \texttt{appears\_in\_review} field for each ablation in the suggested plan.  
   If the ablation is considered a match to one of the reviews, you should set \texttt{appears\_in\_review} to true, otherwise false.

\smallskip
Here are the review(s) of the paper VS. suggested plan:

\smallskip
<official\_review>  
\{\{official\_reviews\}\}  
</official\_review>

\smallskip
<suggested\_plan>  
\{\{problem\_statement\}\}  
</suggested\_plan>

\smallskip
<paper\_source>  
\{\{paper\_source\}\}  
</paper\_source>

\smallskip
BELOW IS AN EXAMPLE THAT WILL HELP YOU UNDERSTAND THE FORMATTING AND THE TASK BETTER.  
PLEASE USE IT FOR REFERENCE ONLY AND DO NOT REPEAT IT IN YOUR FINAL ANSWER.

\smallskip
<official\_review>  
**summary**: This paper proposes EG-SAT, an equivariant graph self-attention transformer for modeling 3D molecular structures, introducing Attention-based Atom-Centered Symmetry Functions (AACSFs) to capture higher-order geometric interactions. The authors showed the performance of proposed model on QM9 and MD17 datasets. While the paper presents some interesting ideas around combining attention mechanisms with ACSFs, there are several critical limitations that need to be addressed.

**soundness**: 1  

**presentation**: 3  

**contribution**: 2  

**strengths**: 1. The integration of attention mechanisms with ACSFs is interesting.  
2. the framework is applicable to multiple molecular property prediction tasks  
3. the paper discusses the theoretical foundation of symmetry and group representation in detail

**weaknesses**: 1. The paper omits several recent works in molecular property prediction, making the comparisons less relevant.  
2. The authors didn't conduct ablation studies to evaluate the contribution of different components in the framework.  
3. There is no computational efficiency analysis, what's more, the claims of improved scalability are unsupported by any experiments.  
4. The motivation for incorporating angular information lacks clear examples where angular information provides benefits.  
5. There's no proof that the attention mechanism preserves chemical validity

**questions**: 1. Can the authors provide experiment for the claimed scalability improvements over traditional ACSFs?  
2. How does the computational complexity scale with the number of atoms and chemical elements compared to existing methods?  
3. What is the memory footprint of the attention mechanism for larger molecular systems?  
4. Can the authors provide ablation studies showing the specific benefits of angular information integration?  
5. How sensitive is the model to hyperparameter choices, particularly the attention and gating parameters?

**flag\_for\_ethics\_review**: ['No ethics review needed.']  

**rating**: 3  

**confidence**: 4  

**code\_of\_conduct**: Yes  
</official\_review>

\smallskip
<discussion>  

The review text contains several weaknesses and questions.  
While in the weaknesses section, the reviewer mentions the lack of ablation studies, it does not specify which ablations are missing.  
However, in the questions section, the reviewer explicitly asks for ablation studies showing the specific benefits of angular information integration.  
The reviewer also suggests the need for experiments to validate the claimed scalability improvements, which could imply a missing ablation, but since we are being conservative we won't consider this an ablation suggestion.

Therefore, the ablation suggestions that can be extracted from the review text are:

1. Angular information integration ablation: The reviewer explicitly requests ablation studies to demonstrate the benefits of incorporating angular information, indicating that this is a missing ablation.

Now let's analyze the ablation in the suggested plan:

1. Remove Radial ACSFs: This ablation focuses on removing the radial component of ACSFs. The review does not mention anything about radial ACSFs, so this is not a match.

2. Remove Angular ACSFs: This ablation involves removing the angular component of ACSFs. This directly corresponds to the reviewer's request for ablation studies on angular information integration.  
   Although the reviewer does not mention the specific action of removal, the ablated part matches exactly, so we consider this a match.

3. Replace Gating Mechanism: This ablation suggests replacing the gating mechanism in ACSFs with alternatives like simple summation or weighted average. The review does not mention anything about the gating mechanism, so this is not a match.

</discussion>

\smallskip
<predictions> 

\{"name\_in\_plan": "Remove Radial AACSFs", "appears\_in\_review": false\}  

\{"name\_in\_plan": "Remove Angular AACSFs", "appears\_in\_review": true\}  

\{"name\_in\_plan": "Replace Gating Mechanism", "appears\_in\_review": false\}  

</predictions>

\end{reviewerpromptbox}
\captionsetup{type=figure}
\captionof{figure}{The task instance prompt for LM judge for \reviewerbench{}.}
}

\subsubsection{\textsc{AgentJudge}}

{
\begin{reviewerpromptbox}{System Prompt}
\textbf{SETTING:} You are an autonomous computer science researcher, an expert at analyzing machine learning papers and their ablation studies, and you're working directly in the command line of a linux container with a special interface.

\smallskip
The special interface consists of a file editor that shows you \{\{WINDOW\}\} lines of a file at a time.  
In addition to typical bash commands, you can also use specific commands to help you navigate within files.  
To call a command, you need to invoke it with a function call/tool call.

\smallskip
\textbf{RESPONSE FORMAT:}  
Your shell prompt is formatted as follows:  

\texttt{(Open file: <path>)}

\texttt{(Current directory: <cwd>)  }

\texttt{bash-\$}

\smallskip
First, you should \_always\_ include a general thought about what you're going to do next.  
Then, for every response, you must include exactly \_ONE\_ tool call/function call.

\smallskip
Remember, you should always include a \_SINGLE\_ tool call/function call and then wait for a response from the shell before continuing with more discussion and commands. Everything you include in the DISCUSSION section will be saved for future reference.  
If you'd like to issue two commands at once, PLEASE DO NOT DO THAT! Please instead first submit just the first tool call, and then after receiving a response you'll be able to issue the second.  
Note that the environment does NOT support interactive session commands (e.g. python, vim), so please do not invoke them.
\smallskip

\end{reviewerpromptbox}
\captionsetup{type=figure}
\captionof{figure}{The system prompt for SWE-agent judge for \reviewerbench{}.} \label{fig:agent_system_prompt_reviewer_judge}
}

{
\begin{reviewerpromptbox}{Instance Prompt}
\textbf{SETTING:} You are in an empty repository root directory \texttt{/repo}.  
We're currently reviewing an ablation studies plan for the research paper \{\{paper\_title\}\}. Here's the research abstract:

\smallskip
\textbf{ABSTRACT:}  
\{\{abstract\}\}

\smallskip
The paper source is provided in the \texttt{/paper} read-only directory.  
You should interpret the review(s) of the paper in the context of the paper source, which may contain important information about the ablation studies performed in the paper, and can help you better understand the ablation suggestions in the review.

\smallskip
\textbf{INSTRUCTIONS:}  
Below, you will find official review(s) of the paper referring to the ablation studies performed in the paper, alongside the suggested ablation plan.  
Your task is to determine, for each ablation in the suggested ablation plan, whether it appears in the review(s).

\smallskip
An ablation in plan is considered to appear in the review if all of the following are true:

1. The \texttt{ablated\_part} (i.e., the component or mechanism being ablated) matches exactly to one of the ablation suggestions in the review text.  

2. The \texttt{action} in one ablation must match exactly (if it exists in the review text ablation suggestion).  

3. If the action is REPLACE or ADD, the \texttt{replacement} content must match (if it exists in the review text ablation suggestion).  
   \smallskip
   * If the field \texttt{"replacement"} contains multiple valid options, your match is valid if at least one correct replacement is identified in the suggested plan.

\smallskip
Be conservative: if it is unclear whether one ablation matches another, do not match.  
However, because the review text may be incomplete, an ablation is still considered a match if the \texttt{ablated\_part} matches exactly (Criterion 1) and the action or replacement information (Criteria 2 and 3) are not mentioned in the review text.

\smallskip
\textbf{What You Need to Do:}  

Create the file \texttt{/repo/final\_score.jsonl} using the command \texttt{create\_final\_score} and then fill in the following fields for each entry:

\smallskip
\{"name\_in\_plan": [ name of the ablation in the plan ], "appears\_in\_review": [ true/false based on the criteria above ]\}

\smallskip
\textbf{GENERAL IMPORTANT TIPS:}

1. The file should be initialized using the command \texttt{create\_final\_score}, after that you just need to edit it and fill in the \texttt{appears\_in\_review} (where applicable) field.  

2. DO NOT modify the field \texttt{name\_in\_plan} at all.  

3. You must go over all the ablations in the suggested plan.  

4. Please note that all review information is in the following text, so you should base your decision only using the information below.  

5. You should treat the review \_ONLY FOR NEW ABLATION SUGGESTIONS\_, and not for the existing ones or other aspects of the paper or the review.  
   For example, the review text may contain suggestions for other experiments which are NOT ablations.  

6. You are given the paper source in the \texttt{/paper} directory — use it to better understand the context of the review and the ablation suggestions.  
   For example, the paper source contains already the ablation studies performed in the paper, which can help you understand better the ablation suggestions in the review.

\smallskip
\textbf{STRATEGY:}

1. First call the \texttt{create\_final\_score} command to create the file.  

2. Then, you should extract from each review the missing ablation that the reviewer suggests to add.  
   A missing ablation refers to an experiment that the reviewer believes should have been conducted, in which a specific component, module, feature, or design choice is removed, replaced, or altered in order to assess its impact on the model's performance.  
   The review must clearly indicate that such an ablation study is missing or should be added.

3. After that, you should have the information to fill in the \texttt{appears\_in\_review} field for each ablation in the suggested plan.  
   If the ablation is considered a match to one of the reviews, you should set \texttt{appears\_in\_review} to true, otherwise false.

4. Submit your final score using the \texttt{submit} command.

\smallskip
PLEASE NOTE THAT THE SUBMISSION FILE NEEDS TO BE INITIALIZED USING THE \texttt{create\_final\_score} COMMAND WHICH INITIALIZES ALL VALUES WITH FALSE AS THE \texttt{appears\_in\_review}, YOU SHOULD CHANGE IT \_FOR EACH\_ OF THE LINES IF THERE IS A VALID MATCH TO THE REVIEW.  
When you are done, run the \texttt{submit} command to confirm.  
You have access to the terminal session, so take your time and be precise.

\smallskip
Here are the review(s) of the paper VS. suggested plan:

\smallskip
<official\_review>  
\{\{official\_reviews\}\}  
</official\_review>

\smallskip
<suggested\_plan>  
\{\{problem\_statement\}\}  
</suggested\_plan>

\smallskip
(Open file: \{\{open\_file\}\})  

(Current directory: \{\{working\_dir\}\})  

\texttt{bash-\$}

\end{reviewerpromptbox}
\captionsetup{type=figure}
\captionof{figure}{The task instance prompt for SWE-agent judge for \reviewerbench{}.}
}

\subsubsection{\textsc{LM-Planner}}

The LM planner system prompt for \reviewerbench{} is identical to the LM planner system prompt for \researcherbench{} given in~\Cref{fig:lm_planner_system_prompt}.

{
\begin{reviewerpromptbox}{Instance Prompt}
We're currently want to suggest missing ablation studies for the research titled \{\{paper\_title\}\}. Here's the research abstract:

\smallskip
ABSTRACT:

\{\{problem\_statement\}\}

\smallskip
The paper source is provided below, after all of the instructions.

\smallskip
INSTRUCTIONS:

Now, you're going to suggest UP TO \{\{num\_ablations\}\} missing ablation studies in the given paper on your own, in a JSONL format.
You need to rank the output ablation studies by their importance, and you should only include the most important ones.
A missing ablation refers to an experiment that you believe should have been conducted, in which a specific component, module, feature, or design choice is removed, replaced, or altered in order to assess its impact on the method's performance.
You must clearly indicate that such an ablation study is missing from the given paper.
Each suggestion should include the following fields in a separate JSON:

1. "name": name of the ablation experiment.

2. "ablated\_part": high-level description of the part of the method in the research that you want to ablate.

3. "action": the action you want to take on the ablated part (REMOVE, REPLACE, ADD). If the action is REPLACE or ADD please add a field named "replacement" and specify a list of possible replacements/additions, for example if you want to change the value of a parameter please specify a list of values to test.

4. "metrics": a list of metrics to report of the ablation experiment. Please pay special attention and use the metrics that are also used in the paper.

When you're satisfied with your ablation studies plan, you can submit your plan.

\smallskip

GENERAL IMPORTANT TIPS:

1. The paper source is provided below, after all of the instructions.

2. Remember that we are looking for \_MISSING IMPORTANT ABLATIONS\_, and as such they should attribute to the method's major components.

3. Less is more - don't aim to change completely the method, take the important parts and investigate them.

\smallskip
STRATEGY:

1. Read the paper sections, especially the method section, to understand the method, and pay attention to the ablation section which is usually under the experiment section or in the appendix.

2. You must first understand and extract the ablation studies that already exist in the paper (use them in your discussion field to better reason about missing ablations).

3. Then, look at the main components of the method and think about what would happen if you change them.

4. Keep in mind that ablation studies aim to attribute the method's performance to its major components, so think about what the ablation studies in the paper are currently missing.

5. Remember that ablation studies does not necessarily mean removing parts of the method, but also changing them.

\smallskip
<paper\_source>

\{\{paper\_source\}\}

</paper\_source>
\end{reviewerpromptbox}
\captionsetup{type=figure}
\captionof{figure}{The task instance prompt for LM planner for \reviewerbench{}.}
}

\subsubsection{\textsc{Agent-Planner}}
The agent planner system prompt for \reviewerbench{} is identical to the agent judge system prompt given in~\Cref{fig:agent_system_prompt}.

{
\begin{reviewerpromptbox}{Instance Prompt}
You are in an empty repository root directory `/repo`. The paper markdown is located in the `/paper` read-only directory.
We're currently want to suggest missing ablation studies for the research titled \{\{paper\_title\}\}. Here's the research abstract:

\smallskip
ABSTRACT:
\{\{problem\_statement\}\}

\smallskip
INSTRUCTIONS:

Now, you're going to suggest UP TO \{\{num\_ablations\}\} missing ablation studies in the given paper on your own, in a JSONL file `/repo/ablations\_plan.jsonl`.
You need to rank the output ablation studies by their importance, and you should only include the most important ones.
A missing ablation refers to an experiment that you believe should have been conducted, in which a specific component, module, feature, or design choice is removed, replaced, or altered in order to assess its impact on the method's performance.
You must clearly indicate that such an ablation study is missing from the given paper.
Each suggestion should include the following fields in a separate JSON:

\smallskip
1. "name": name of the ablation experiment.

2. "ablated\_part": high-level description of the part of the method in the research that you want to ablate.

3. "action": the action you want to take on the ablated part (REMOVE, REPLACE, ADD). If the action is REPLACE or ADD please add a field named "replacement" and specify a list of possible replacements/additions, for example if you want to change the value of a parameter please specify a list of values to test.

4. "metrics": a list of metrics to report of the ablation experiment. Please pay special attention and use the metrics that are also used in the paper.

\smallskip
Your terminal session has started and you're in the repository's root directory.
You can use any bash commands or the special interface to help you.
The environment is already configured. YOU DON'T NEED TO INSTALL ANYTHING IN THE ENVIRONMENT.
Remember, YOU SHOULD ALWAYS INCLUDE EXACTLY ONE TOOL CALL/FUNCTION CALL PER RESPONSE.
When you're satisfied with your ablation studies plan, you can submit your plan to the code base by simply running the submit command.
Note however that you cannot use any interactive session commands (e.g. python, vim) in this environment, but you can write scripts and run them. E.g. you can write a python script and then run it with the python command.

\smallskip
GENERAL IMPORTANT TIPS:

1. To view the paper you should first run `cd /paper`.

2. Remember that we are looking for \_MISSING IMPORTANT ABLATIONS\_, and as such they should attribute to the method's major components.

3. Less is more - don't aim to change completely the method, take the important parts and investigate them.

4. Don't try to run extensive operations, remember that we just want to get the missing ablations plan for now.

5. If you run a command and it doesn't work, try running a different command. A command that did not work once will not work the second time unless you modify it!

6. Please note that in order to open a file that contains spaces you MUST use single quotes like this: `open 'file that contains spaces.md'`. YOU SHOULD NOT USE ANY DOUBLE QUOTES TO OPEN THE FILE AT ALL.

7. If you open a file and need to get to an area around a specific line that is not in the first 100 lines, say line 583, don't just use the scroll\_down command multiple times. Instead, use the goto 583 command. It's much quicker.

8. Always make sure to look at the currently open file and the current working directory (which appears right after the currently open file). The currently open file might be in a different directory than the working directory! Note that some commands, such as 'create', open files, so they might change the current open file.

\smallskip
STRATEGY:

1. Read the paper sections, especially the method section, to understand the method, and pay attention to the ablation section which is usually under the experiment section or in the appendix.

2. You must first understand and extract the ablation studies that already exist in the paper (don't save them to a file, just use them in your process to generate new ablations).

3. Then, look at the main components of the method and think about what would happen if you change them.

4. Keep in mind that ablation studies aim to attribute the method's performance to its major components, so think about what the ablation studies in the paper are currently missing.

5. Remember that ablation studies does not necessarily mean removing parts of the method, but also changing them.

\smallskip
(Open file: \{\{open\_file\}\})

(Current directory: \{\{working\_dir\}\})

bash-\$
\end{reviewerpromptbox}
\captionsetup{type=figure}
\captionof{figure}{The task instance prompt for LM planner for \reviewerbench{}.}
}

\section{AI Assistant Usage}
\label{app:ai_usage}

AI assistants were used solely for coding assistance (GitHub Copilot) and for grammatical corrections and sentence rephrasing during the writing process (ChatGPT). AI tools were not used for the ideation of this research project, the creation of the dataset, or any other decisions regarding the research directions.